%% file: main.tex
\documentclass{article}

\usepackage[preprint]{neurips_2026}

\usepackage[utf8]{inputenc} %
\usepackage[T1]{fontenc}    %
\usepackage{hyperref}       %
\usepackage{url}            %
\usepackage{booktabs}       %
\usepackage{amsfonts}       %
\usepackage{nicefrac}       %
\usepackage{microtype}      %
\usepackage{xcolor}         %

\usepackage{graphicx}
\usepackage{amsmath}
\usepackage{cleveref}
\usepackage{tikz}
\usepackage{multirow}
\usepackage{array} 
\usepackage{makecell}
\usepackage{booktabs}
\usetikzlibrary{spy}
\usepackage{subcaption}

\crefname{figure}{Figure}{Figures}
\crefname{table}{Table}{Tables}
\crefname{section}{Section}{Sections}
\Crefname{section}{Section}{Sections}
\crefname{subsection}{Section}{Sections}
\crefname{subsubsection}{Section}{Sections}
\crefname{appendix}{Appendix}{Appendices}
\usepackage{enumitem}
\usepackage{placeins}
\newcommand{\first}[1]{\textbf{#1}}      %
\newcommand{\second}[1]{\underline{#1}}  %
\title{OPERA: An Agent for Image Restoration with End-to-End Joint Planning–Execution Optimization
}

\author{%
{Feng Zhu$^{1}$\thanks{Equal contribution. $\dag$Corresponding author.} ~ ~ ~ Shuyang Xie$^{1*}$ ~ ~ ~ Yihan Zeng{$^{2}$} ~ ~ ~ Ming Liu{$^{1}$} ~ ~ ~ Wangmeng Zuo{$^{1\dag}$} }\\
\normalsize
\vspace{2pt}
$^{1}$\	Harbin Institute of Technology  \quad
$^{2}$\ Huawei Noah’s Ark Lab
}

\begin{document}

\maketitle

\input{sections/0_abstract}

\input{sections/1_introduction}

\input{sections/2_related_work}

\input{sections/2_5_Analysis}

\input{sections/3_method}

\input{sections/4_experiments}

\input{sections/5_conclusion}

\bibliographystyle{plain}
\bibliography{main}

\appendix

\input{sections/6_appendix}

\FloatBarrier

\end{document}

%% file: sections/0_abstract.tex
\begin{abstract}
Real-world image restoration is challenging due to complex and interacting mixed degradations.
Recent agent-based approaches address this problem by composing multiple task-specific restoration tools.
However, empirical analysis reveals that their performance is fundamentally limited by implicitly constrained planning spaces and the lack of coordination among independently pretrained tools.
To address these issues, we propose \textbf{OPERA} (Optimized Planning-Execution Restoration Agent), a framework that jointly optimizes restoration planning and tool execution in an end-to-end manner.
On the planning side, OPERA uses reinforcement learning to directly optimize tool composition over a combinatorial plan space, with the final restoration quality as the reward.
On the execution side, OPERA introduces agent-guided co-training of restoration tools, enabling them to learn cooperative behaviors under sequential composition.
Extensive experiments on multi-degradation benchmarks and real-world datasets demonstrate that OPERA consistently outperforms both all-in-one restoration models and existing agent-based methods across diverse and complex degradation scenarios.
Codes are available at: \url{https://github.com/xsyshuishui/Opera}.

\end{abstract}

%% file: sections/1_introduction.tex
\section{Introduction}
\label{sec:intro}
Real-world images often suffer from complex degradations involving multiple distortion types simultaneously, where noise, blur, haze, rain, and compression artifacts can coexist and interact in non-trivial ways. Unlike the clean, single-degradation scenarios commonly studied in benchmarks, mixed degradations pose significant challenges for image restoration systems, as the combined effects of multiple distortions cannot be modeled simply.

Early efforts address this problem by developing all-in-one restoration models that aim to handle multiple degradation types within a single network~\cite{zamir2022restormer, Promptir, airnet, daclip}. While conceptually appealing, these models inevitably face a trade-off between generalization and specialization: accommodating diverse degradation patterns often results in overly smooth outputs and the loss of fine details, especially in complex real-world settings~\cite{chen2024restoreagent}.

Recently, agent-based image restoration methods have emerged as a promising alternative~\cite{agenticir,chen2024restoreagent}. Instead of relying on a single model, these approaches leverage a collection of off-the-shelf, task-specific restoration tools to restore the image. By leveraging large language models (LLMs) or vision-language models (VLMs), they design agentic systems to dynamically select and compose appropriate tools for a given degraded input. By exploiting the strengths of specialized models, agentic systems have the potential to achieve strong task-specific performance while handling arbitrary combinations of degradations. This paradigm has opened a new direction for handling complex real-world degradations beyond the scope of conventional single-model designs.

Despite this, existing agent-based methods suffer from two fundamental limitations that hinder their effectiveness in complex multi-degradation scenarios.
\begin{itemize}[leftmargin=1em]
\item\textbf{(1) Implicitly constrained planning space.}
Existing agent-based image restoration methods rely on implicit planning assumptions that significantly restrict the space of restoration plans. For example, many methods assume a one-to-one mapping between degradation types and restoration tools, planning by explicitly matching each detected degradation to a specific tool~\cite{agenticir, chen2024restoreagent}. Some methods plan in a stepwise greedy manner, where each action is selected to locally improve an intermediate quality metric~\cite{lu2025simplecall,zhou2025q}. These assumptions substantially narrow the planning space, limiting the agent’s ability to discover more complex yet effective cooperative restoration strategies.

\item\textbf{(2) Static pretrained tools without coordination.}
Existing agent-based frameworks treat restoration tools as fixed, independently pretrained modules~\cite{agenticir,chen2024restoreagent}. When these tools are composed sequentially, the output distribution of one tool becomes the input to the next, but these tools were never trained to cooperate. Applying a tool may alter the distribution of remaining degradations, adversely affecting subsequent restoration steps. Prior work has attempted to alleviate this issue by searching for optimal tool orderings, but we argue that ordering alone is insufficient. When tools lack coordinated training, no single ordering can consistently yield satisfactory results across diverse degradation combinations.
\end{itemize}
To concretely understand how these limitations affect performance, we conduct an empirical study on cooperative multi-tool image restoration in a constrained setting. By exhaustively enumerating restoration plans, we analyze which tool combinations yield high-quality restoration outcomes. As illustrated in~\cref{fig:empirical}, many high-performing plans violate common planning assumptions. For example, effective restoration often involves out-of-scope tools as well as repeated tool applications. This also demonstrates the limitations of existing restoration tools when applied to this cooperative setting. The findings suggest that both a more expressive planning space and improved coordination among tools are necessary for effective agent-based restoration.

Motivated by these observations, we propose \textbf{OPERA}, a framework that jointly optimizes restoration planning and tool execution in an end-to-end manner. At the \emph{planning} level, OPERA departs from a hand-crafted, step-by-step decision-making workflow and instead trains an agent to generate a complete tool-invocation plan end-to-end. Given the combinatorial nature of the tool composition space, we use reinforcement learning to optimize the agent, with the final restoration quality serving as the reward signal. This formulation enables the agent to reason globally over tool compositions and discover non-obvious combinations.
At the \emph{execution} side, OPERA proposes agent-guided tool model training. Rather than treating restoration tools as static, independently pretrained modules, we jointly fine-tune tool models under the agent's generated plans. In this process, the agent acts as a high-level planner that induces diverse tool compositions, while individual tools remain architecturally independent and are updated solely based on their contribution to downstream restoration quality. This co-training strategy enables tools to learn cooperative behaviors, effectively mitigating the distribution shift caused by sequential composition.

Extensive experiments demonstrate that OPERA significantly outperforms both all-in-one restoration models and existing agent-based systems across most metrics. We further conduct analysis showing that the agent learns non-trivial planning strategies, while jointly trained tools adapt to operate robustly within agent-generated restoration plans. Finally, OPERA generalizes well to real-world datasets, highlighting the robustness of our joint planning-and-execution framework.

Our main contributions are summarized as follows:
\begin{itemize}[leftmargin=2em]
    \item We present an empirical study of cooperative multi-tool image restoration, providing key insights into the design of the agentic image restoration system.
    \item We propose \textbf{OPERA}, an end-to-end agent-based restoration framework that jointly optimizes tool composition planning via reinforcement learning and enables cooperative behavior through agent-guided tool training.
    \item Extensive experiments on multi-degradation benchmark and real-world dataset show that OPERA significantly outperforms existing all-in-one models and agent-based methods across most metrics.
\end{itemize}

%% file: sections/2_related_work.tex
\section{Related Work}
\label{sec:related}

\subsection{All-in-One Image Restoration}

All-in-One Image Restoration (AiOIR) aims to develop unified models that handle diverse degradation types within a single framework~\cite{jiang2025survey}. Early approaches adopt shared encoder-decoder architectures with multi-scale processing~\cite{zamir2022restormer}. Recent advances explore various conditioning mechanisms: prompt-based methods~\cite{Promptir} use learnable prompts for task adaptation, degradation embedding approaches~\cite{airnet,daclip} explicitly encode degradation representations, and Mixture-of-Experts architectures route inputs to specialized sub-networks. While these methods achieve strong performance on standard benchmarks, they may struggle with complex real-world degradation mixtures~\cite{lu2025simplecall} and face trade-offs between generalization and task-specific accuracy~\cite{chen2024restoreagent}. Moreover, incorporating new degradation types typically requires retraining the entire model, limiting extensibility~\cite{agenticir}.

\subsection{Agent-based Image Restoration}

Agent-based methods address AiOIR limitations by employing intelligent controllers that dynamically select and sequence restoration tools. We review existing approaches along two dimensions:%

\paragraph{Planning Strategies.}
Existing agent-based methods universally adopt step-by-step planning with iterative execution. RestoreAgent~\cite{chen2024restoreagent} first leverages multimodal LLMs for degradation identification, introducing iterative replanning with rollback mechanisms to correct suboptimal decisions. AgenticIR~\cite{agenticir} systematizes the process into a five-stage perception-scheduling-execution-reflection-rescheduling pipeline, while Q-Agent~\cite{zhou2025q} incorporates quality-driven Chain-of-Thought~\cite{wei2022chain} reasoning to guide tool selection. To improve efficiency, SimpleCall~\cite{lu2025simplecall} replaces heavy LLM inference with a lightweight policy network trained via PPO~\cite{schulman2017proximal}, achieving label-free learning but still following the iterative state-action-state paradigm. Despite their diversity, these methods share implicit assumptions that constrain the planning space. Degradation-matching approaches (RestoreAgent, AgenticIR) assume a one-to-one correspondence between detected degradations and applicable tools, while step-by-step methods (Q-Agent, SimpleCall) select tools based on immediate quality improvement. However, as we demonstrate in~\cref{sec:empirical}, high-performing restoration plans frequently violate these assumptions: effective restoration often involves out-of-scope tools and repeated tool applications that cannot be discovered through local optimization. In contrast, our agent generates complete tool plans in a single forward pass and is trained end-to-end to optimize final restoration quality directly.

\paragraph{Tool Utilization.}
Early RL-based work RLRestore~\cite{yu2018crafting} assembles small CNN-based tools trained independently for specific degradations. Subsequent LLM-based approaches~\cite{chen2024restoreagent,agenticir,zhou2025q,lu2025simplecall} inherit this practice, relying on off-the-shelf restoration models without adaptation. 4KAgent~\cite{zuo20254kagent} attempts to improve tool selection via Mixture-of-Experts routing, but its experts remain independently trained. Despite varied architectures, these methods universally treat tools as fixed modules. In contrast, our work jointly fine-tunes tools under agent-generated plans.

%% file: sections/2_5_Analysis.tex
\section{Cooperative Multi-Tool Image Restoration}
\label{sec:empirical}

As discussed in~\cref{sec:intro}, recent agentic image restoration approaches typically rely on a collection of task-specific restoration tools and restore a degraded image by sequentially composing these tools. While this paradigm has been widely adopted, prior work often relies on implicit, largely unverified assumptions about how restoration tools should be composed.
In this section, we first formalize the problem of cooperative image restoration with multiple tools. Then, through a controlled empirical study in a constrained setting where the action space can be exhaustively explored, we analyze the characteristics of high-performing restoration plans.

\begin{figure}[!tb]

  \begin{center}
    \centerline{\includegraphics[width=\columnwidth]{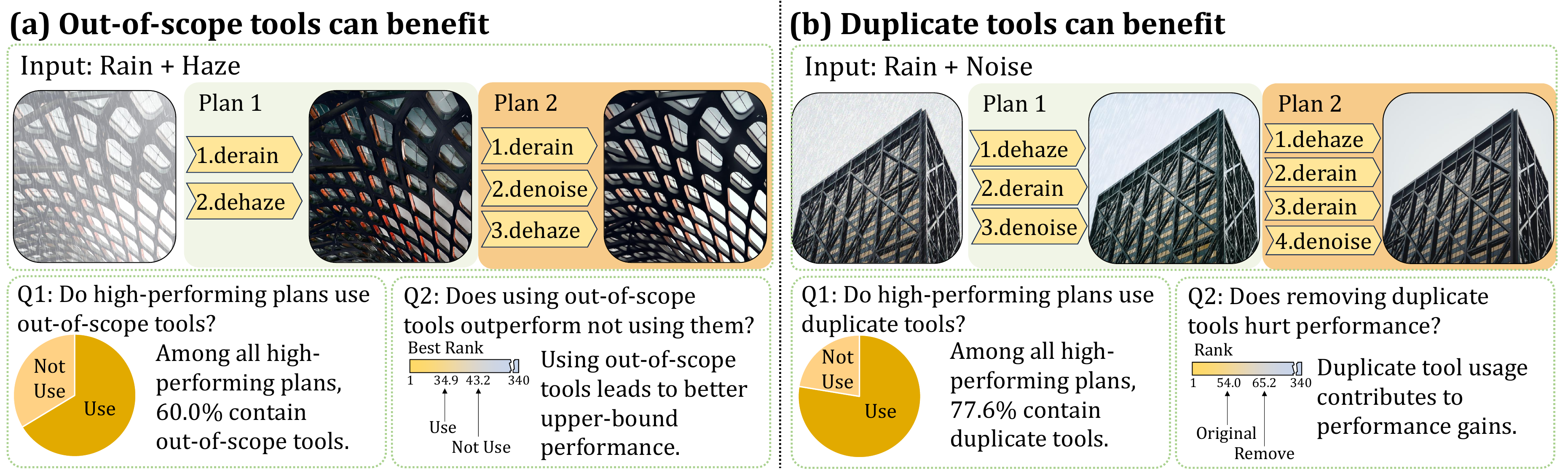}}
    \caption{
Empirical study of cooperative multi-tool image restoration. Zoom in to see image details.
    }
    \label{fig:empirical}
  \end{center}
  \vskip -0.25in
\end{figure}

\subsection{Problem Formulation}
Assume we have collected a library of task-specific restoration tools $\mathcal{M}$, where each tool is trained to target a specific type of degradation. Given a degraded input image $I_{\mathrm{LQ}}$, the goal is to design a restoration plan $\mathcal{P} = [M_1,M_2,\ldots,M_k]$, where each element $M_i \in \mathcal{M}$. By sequentially applying these tools to $I_{\mathrm{LQ}}$, we obtain the restored image $I_{\mathrm{pred}}$. The quality of a restoration plan is evaluated by image quality assessment metrics applied to $I_{\mathrm{pred}}$.

\subsection{Empirical Analysis via Exhaustive Search}

To gain insight into what constitutes an effective tool composition, we conduct controlled empirical experiments in which the space of restoration plans can be exhaustively enumerated. Our objective is to empirically analyze the characteristics of high-quality restoration plans in a multi-tool cooperative setting.

\textbf{Experimental Settings.}
To make exhaustive search tractable, we consider a reduced setting with four common degradations: noise, rain, haze, and blur, each associated with a dedicated restoration tool. Restoration quality is evaluated using both full-reference metrics PSNR, SSIM~\cite{SSIM}, LPIPS~\cite{LPIPS}, and no-reference metrics CLIP-IQA~\cite{CLIPIQA}, MUSIQ~\cite{MUSIQ}, which capture complementary aspects of image fidelity and perceptual quality.

We randomly select 15 images from the MiOIR-Test~\cite{kong2024towards} set as clean high-quality images, and synthesize low-quality inputs by applying 8 manually designed combinations of degradations, yielding 120 degraded images. For each degraded input, we enumerate all possible restoration plans with a maximum length of 4, where each step selects one of 4 available tools, yielding a total of 340 candidate plans per input. Details are shown in~\cref{appendix:empirical}.

\textbf{Selection of High-Performing Plans.}
After evaluating all candidate plans, we rank them independently for each IQA metric. For a given metric, a plan is considered \emph{good} if it ranks within the top 10$\%$ among all plans. To identify plans that perform robustly across different quality criteria, we select those ranked \emph{good} by at least 3 of the 5 metrics, with the additional requirement that both full-reference and no-reference metrics are included. This selection strategy mitigates metric-specific bias and emphasizes consistent performance across complementary evaluation signals.
The resulting set of high-performing plans forms the basis of our subsequent analysis. On average, 12.05 out of the 340 plans per image are selected as high-performing. In addition, we compute an aggregated rank score for each plan by averaging its rank across all five metrics.

\textbf{Finding 1: Out-of-Scope Tools Can Be Beneficial.}
Since the low-quality inputs are synthetically generated, the underlying degradation types are known for each image. Interestingly, despite this knowledge, we observe that a substantial portion of high-performing plans include restoration tools that do not correspond to any of the degradations present in the input. Specifically, 60.0$\%$ of the selected high-performing plans contain at least one out-of-scope tool. As illustrated in ~\cref{fig:empirical} (a), applying a denoising tool can improve restoration quality even when the input image is not corrupted by noise in some cases.
To further quantify this effect, we compare, for each input image, the best aggregated rank achieved by plans that include out-of-scope tools with that of plans that strictly match the ground-truth degradation types. Plans incorporating out-of-scope tools outperform matched-only plans on 66.3$\%$ of the images, achieving a lower (better) mean best rank of 34.9 compared to 43.2.

\textbf{Finding 2: Duplicate Tools Can be Beneficial.}
We further observe that repeated application of the same tool is common among high-performing plans: 77.6$\%$ of high-performing plans apply at least one tool multiple times. To assess whether such duplication meaningfully contributes to restoration quality, we compare each high-performing plan containing duplicate tools with its de-duplicated counterpart. As shown in ~\cref{fig:empirical} (b), de-duplication consistently degrades performance, increasing the average aggregated rank from $54.0$ to $65.3$.
While a duplicate application is not universally beneficial, it can play a critical role in high-performing restoration plans.

\textbf{Analysis.}
Together, these findings indicate that effective cooperative image restoration plans often deviate from intuitive heuristics such as matching tools to known degradations. Instead, complex tool compositions can outperform simpler, more constrained strategies.
More broadly, our results highlight the limitations of existing single-degradation restoration tools when applied to cooperative settings. Models trained in isolation for specific degradations may require careful orchestration or iterative application. These observations motivate the co-evolution of the planning agent and the restoration tools.

%% file: sections/3_method.tex
\section{Method}
\label{sec:method}

\begin{figure}[t]
  \centering
  \includegraphics[width=\columnwidth]{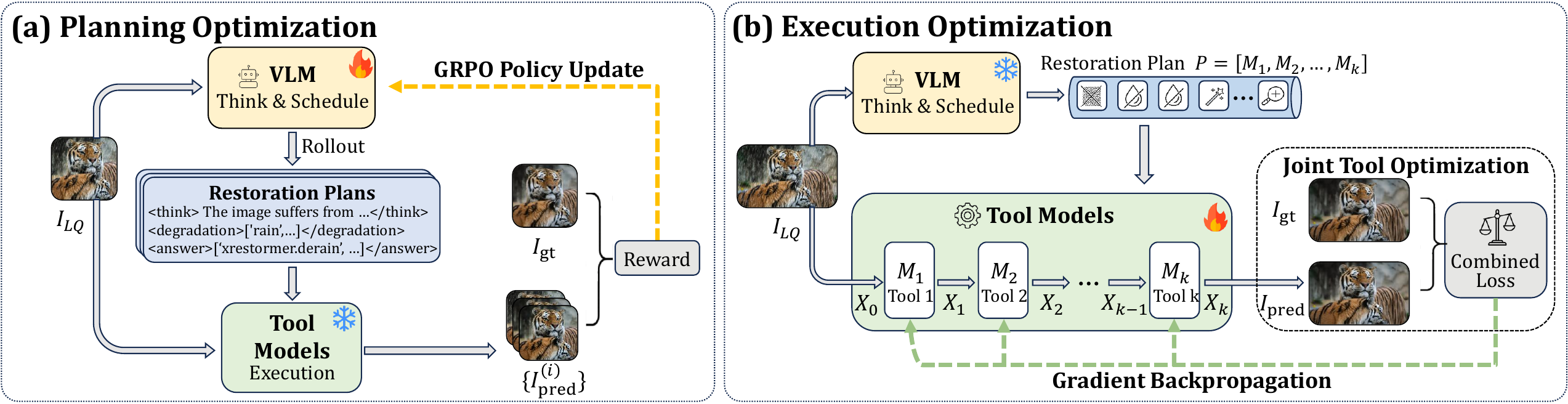}
  \caption{Overview of our OPERA framework.
    (a) \textbf{Planning Optimization}: The restoration agent is trained via Group Relative Policy Optimization (GRPO) to generate complete restoration plans end-to-end, receiving rewards based on final image quality.
    (b) \textbf{Execution Optimization}: At inference time, the agent generates a restoration plan that is executed by specialized tools. The tools are jointly optimized under agent guidance to cooperate effectively in multi-degradation scenarios.}
  \label{fig:pipeline}
\vskip -0.2in

\end{figure}

In this section, we present the proposed OPERA framework. We begin by describing the overall inference workflow in \cref{method_overall_workflow}. We then detail the OPERA framework, including the planning optimization (\cref{method_agent_train}) and the execution optimization (\cref{method_tool_train}), as illustrated in \cref{fig:pipeline}.

\subsection{Overall Workflow}
\label{method_overall_workflow}
To approximate real-world image restoration scenarios, we consider a multi-degradation image restoration setting.
Following prior works~\cite{chen2024restoreagent, agenticir, zuo20254kagent}, we focus on eight commonly observed degradations: low resolution, noise, motion blur, defocus blur, rain, haze, JPEG compression artifacts, and low light.
For each degradation, a set of specialized restoration tools is collected.
Given a degraded input image, the objective of the restoration agent is to identify an optimal sequence of tool invocations that progressively restores the image toward a clean version. 

The inference pipeline is illustrated in \cref{fig:pipeline}. Given a degraded image, the planning agent generates a restoration plan in a single forward pass, specifying both the selected tools and their execution order. The restoration tools are then sequentially applied to the input image according to the plan, producing the final restored output.

\subsection{Planning Optimization}
\label{method_agent_train}

As discussed in~\cref{sec:intro}, existing agent-based image restoration methods often rely on carefully designed workflows or constrained planning strategies. While effective in specific cases, such designs inherently restrict the space of admissible restoration plans and limit the agent’s ability to discover better tool compositions, as validated in~\cref{sec:empirical}.
In this work, we remove such hand-crafted or step-by-step decision-making workflow, and instead enable the agent to plan freely over the full combinatorial space of tool sequences.
To support this objective, the agent must possess both strong degradation perception and prior knowledge of restoration tool behaviors and their interactions. We therefore initialize the agent with a pretrained vision-language model exhibiting strong image quality assessment capabilities, providing a robust prior for degradation understanding. We further optimize this model to function as an effective restoration planner.

Importantly, for a given degraded image, there is typically no unique or well-defined “ground-truth” restoration plan, even among human experts, and the combinatorial space of possible tool compositions is prohibitively large. These characteristics make supervised learning or manually designed planning heuristics impractical.
Consequently, we adopt reinforcement learning (RL) to optimize the agent in an end-to-end manner. We choose group relative policy optimization (GRPO)~\cite{shao2024deepseekmath} as the RL algorithm. By using the final restoration quality as the reward signal, the agent is encouraged to reason globally across tool compositions and autonomously discover effective, potentially non-obvious restoration strategies.

Notably, we adopt a single-shot inference strategy, in which the agent generates a complete restoration plan in a single forward pass conditioned on the degraded image, rather than relying on multi-turn iterative planning with intermediate tool execution. This design is motivated by both training efficiency and the characteristics of the task. Unlike tasks where step-by-step feedback is beneficial, restoration steps that appear locally optimal do not always align with the global objective. 
By generating the entire plan upfront, the agent is encouraged to perform global reasoning over tool interactions, thereby avoiding suboptimal decisions that may arise from sequential planning.

\textbf{Reward Design.} We adopt an end-to-end reward function to facilitate GRPO training. A restoration plan is considered effective if it leads to a high-quality restored image. Therefore, the primary reward is derived from Image Quality Assessment (IQA) metrics, serving as a proxy for perceptual quality of the restored image. We additionally introduce auxiliary rewards for degradation prediction, structured output formatting, and reasoning–action consistency. The reward consists of four components:
\begin{itemize}[leftmargin=2em]
    \item Restoration Reward $R_q\in \mathbb{R}^+$: This reward evaluates the quality of the restoration plan produced by the agent.
    Specifically, the restoration tools specified in the plan are applied to the input image, and the resulting image is assessed using image quality assessment (IQA) metrics.
    In practice, we adopt a composite objective combining full-reference and no-reference IQA metrics, as a balanced proxy for perceptual quality, following prior works~\cite{chen2024restoreagent, zhou2025q}.
    
    \item Degradation Prediction Reward $R_d\in [0,1]$: This reward measures whether the model correctly predicts the degradations present in the input image. It is computed as the standard $F1$ score between the predicted degradation set and the ground-truth set. Since degradations are synthetically applied during data generation, the ground-truth degradation labels are available.
    We introduce this auxiliary reward to facilitate RL exploration. Degradation prediction is simpler than plan generation, yet accurate degradation identification naturally informs the restoration strategy. By providing explicit supervision on this intermediate subtask, $R_d$ guides the agent toward a better understanding of image conditions and enables more efficient discovery of effective restoration plans during GRPO training.
    \item Format Reward $R_f\in\{0,1\}$: The format reward enforces a structured output that enables reliable parsing and tool execution. The model is instructed to think first, then predict a degradation set, and finally output a restoration plan.
    \item Consistency Reward $R_c\in\{0,1\}$: Following prior works~\cite{zhang2025r1,team2025kwai}, we add this reward that evaluates the alignment between the reasoning process and the final restoration plan, determined by an LLM judge.
    This reward encourages effective reasoning and prevents the model from collapsing into non-thinking behaviors.

\end{itemize}

The final reward is obtained by aggregating the 4 rewards: $R = R_q\times R_d\times R_f \times R_c$.

\subsection{Execution Optimization}
\label{method_tool_train}

While the trained agent learns to generate effective restoration plans, the quality of the final output depends equally on the tools themselves.
Off-the-shelf restoration tools are typically optimized for single-degradation scenarios in isolation.
However, our analysis in~\cref{sec:empirical} highlights the limitations of existing single-degradation restoration tools, and a further optimization is needed for this cooperative setting.
As illustrated in \cref{fig:pipeline} (b), we propose to jointly optimize the restoration tools under the agent's guidance.

\textbf{End-to-End Tool Training.}
Given a restoration plan $\mathcal{P} = [M_1, M_2, \ldots, M_K]$ generated by the agent, the tools are applied sequentially:
\begin{equation}
x_0 = I_{\mathrm{LQ}}, \quad x_k = M_k(x_{k-1}), \quad I_{\mathrm{pred}} = x_K
\end{equation}
The training loss is computed directly between the final output $I_{\mathrm{pred}}$ and the ground truth $I_{\mathrm{gt}}$, with gradients propagating through the entire tool chain.
This end-to-end formulation encourages tools to cooperate rather than optimize independently, as each tool's parameters are updated based on the quality of the final restored image rather than intermediate results.

\textbf{Training Objective.}
We employ a composite loss that balances multiple objectives:
\begin{equation}
\begin{aligned}
\mathcal{L} = {} & w_{\text{pixel}} \mathcal{L}_{\text{L1}} + w_{\text{perc}} \mathcal{L}_{\text{VGG}} + w_{\text{lpips}} \mathcal{L}_{\text{LPIPS}}  + w_{\text{nr}} (\mathcal{L}_{\text{MUSIQ}} + \mathcal{L}_{\text{CLIPIQA}})
\end{aligned}
\end{equation}
where pixel-level loss ($\mathcal{L}_{\text{L1}}$) ensures fidelity, perceptual losses ($\mathcal{L}_{\text{VGG}}$, $\mathcal{L}_{\text{LPIPS}}$) capture visual similarity, and no-reference quality losses ($\mathcal{L}_{\text{MUSIQ}}$, $\mathcal{L}_{\text{CLIPIQA}}$) promote realistic appearance.
To ensure stable training, we employ a progressive loss schedule that transitions from pixel-level to perceptual optimization.
Implementation details and complete loss formulations are provided in~\cref{app:tool_training}.

%% file: sections/4_experiments.tex
\section{Experiments}

This section presents a comprehensive experimental evaluation of the proposed OPREA framework. We compare our framework with state-of-the-art all-in-one restoration models and recent agent-based systems on standard multi-degradation benchmarks, reporting both quantitative and qualitative results. We analyze the planning behavior learned by the agent in~\cref{exp:planning}. We further evaluate generalization performance on real-world degraded images in~\cref{exp:real_world}.
Finally, we analyze the efficiency of our framework in~\cref{exp:efficiency}.

\subsection{Experimental Setting}
\label{exp:settings}
\input{tables/main_metric}

\textbf{Training Data.}
To train both the agent and restoration tools, we require a dataset of multi-degradation images paired with high-quality ground truth. Following AgenticIR~\cite{agenticir}, we use MiOIR-Train~\cite{kong2024towards} as the source of high-quality images and apply different combinations of degradations to synthesize low-quality inputs. We design 145 degradation combinations, with each image containing up to 3 degradations. To reflect realistic scenarios, we only retain combinations that are likely to occur in practice. During training, we sample degradation combinations with a ratio of single:dual:triple=1:3:5, emphasizing complex multi-degradation scenarios. This process yields approximately 16,000 training images.%

\textbf{Restoration Tools.} We select 16 task-specific pretrained tools from Restormer~\cite{zamir2022restormer}, X-Restormer~\cite{chen2024comparative}, and SwinIR~\cite{liang2021swinir} to form the tool pool. These tools cover common degradation types, including noise, blur, rain, haze, and low resolution. Detailed tool specifications are provided in~\cref{appendix:tools}. Note that the chosen tool set is a subset of those of baseline agentic methods.

\textbf{Metrics.} For quantitative evaluation, we adopt both full-reference metrics PSNR, SSIM~\cite{SSIM}, LPIPS~\cite{LPIPS}, and no-reference metrics MANIQA~\cite{MANIQA}, CLIP-IQA~\cite{CLIPIQA}, MUSIQ~\cite{MUSIQ}.  PSNR and SSIM are computed on the Y channel in YCbCr, following 4KAgent~\cite{zuo20254kagent}. Full-reference metrics measure pixel-level fidelity against ground truth, while no-reference metrics assess perceptual quality.

\textbf{Implementation Details.}
For planning optimization, we adopt VisualQualityR1~\cite{wu2025visualquality} as the base model, a finetuned variant of Qwen2.5-VL-7B-Instruct~\cite{bai2025qwen2} equipped with basic image quality assessment capabilities. We adopt Pangu-Embedded-7B~\cite{chen2025pangu} as the LLM judge to calculate consistency reward, due to its strong reasoning capability and high inference throughput. Prompts are detailed in~\cref{app:prompt}. We employ verl~\cite{sheng2024hybridflow} for GRPO training with a batch size of 32 and a group size of $G=8$. The restoration reward $R_q$ is defined as a weighted sum of five IQA metrics: PSNR, SSIM, LPIPS, CLIP-IQA, and MUSIQ.
For tool training, we use the Adam optimizer with a learning rate of $1 \times 10^{-6}$. The training loss combines L1, VGG perceptual, LPIPS, MUSIQ, and CLIP-IQA with weights 0.4, 0.1, 0.15, 0.1, 0.1, respectively. We employ a progressive schedule that transitions from pixel-level to perceptual optimization over the first 30\% of training. All tool backbones (Restormer, X-Restormer, SwinIR) are fine-tuned with a low learning rate to adapt to the agent's calling patterns.

\textbf{Benchmarks.}
We follow the evaluation protocol of AgenticIR~\cite{agenticir} and 4KAgent~\cite{zuo20254kagent}, using the same benchmark images: Groups A, B, and C. These three test sets contain 1,440 LQ images processed with 16 combinations of mixed 2 or 3 types of degradations applied to images from MiOIR-Test~\cite{kong2024towards}.

\subsection{Main Results}

\textbf{Baselines.}
We compare our method against two categories of approaches: (1) all-in-one restoration models, including AirNet~\cite{airnet}, PromptIR~\cite{Promptir}, MiOIR~\cite{kong2024towards}, DA-CLIP~\cite{daclip}, InstructIR~\cite{Instructir}, and AutoDIR~\cite{Autodir}; and (2) agent-based restoration systems, including AgenticIR~\cite{agenticir}, MAIR~\cite{jiang2025multi}, and 4KAgent~\cite{zuo20254kagent}.

\textbf{Quantitative Comparison.}
\cref{table:quantitative_comparison} reports quantitative results for two variants of our approach. ``Ours (Planning)'' corresponds to our variant in which only planning is optimized, while using fixed, pretrained tools. ``Ours (Full)'' is the full version where both planning and execution are optimized.

First, even without tool optimization, the planning-only variant achieves performance on par with or even surpassing existing agentic systems, demonstrating that end-to-end optimization over complete tool composition plans is more effective than heuristic search or greedy-based planning. This highlights the advantage of learning a global restoration strategy directly from the quality of the final image as feedback.
When tool training is enabled, performance improves substantially across all metrics and degradation groups. In Group C, the most challenging setting with complex degradation combinations, the full system achieves improvements of +2.29~dB in PSNR and -0.17 in LPIPS compared to the planning-only variant. This confirms that coordinated tool training is critical for robust restoration under complex degradation scenarios.

\begin{figure*}[t]
\centering
\setlength{\tabcolsep}{1pt}
\newcommand{\zoombox}[4]{%
\begin{tikzpicture}[
    spy using outlines={rectangle, #2, magnification=2, size=1.4cm, connect spies}
]
    \node[anchor=south west,inner sep=0] (img) at (0,0) {\includegraphics[width=0.19\textwidth]{#1}};
    \spy on (#3,#4) in node [anchor=north east] at (img.north east);
\end{tikzpicture}%
}
\begin{tabular}{ccccc}
\multicolumn{5}{c}{\small\textit{Motion Blur + Defocus Blur + Noise}} \\[2pt]
\zoombox{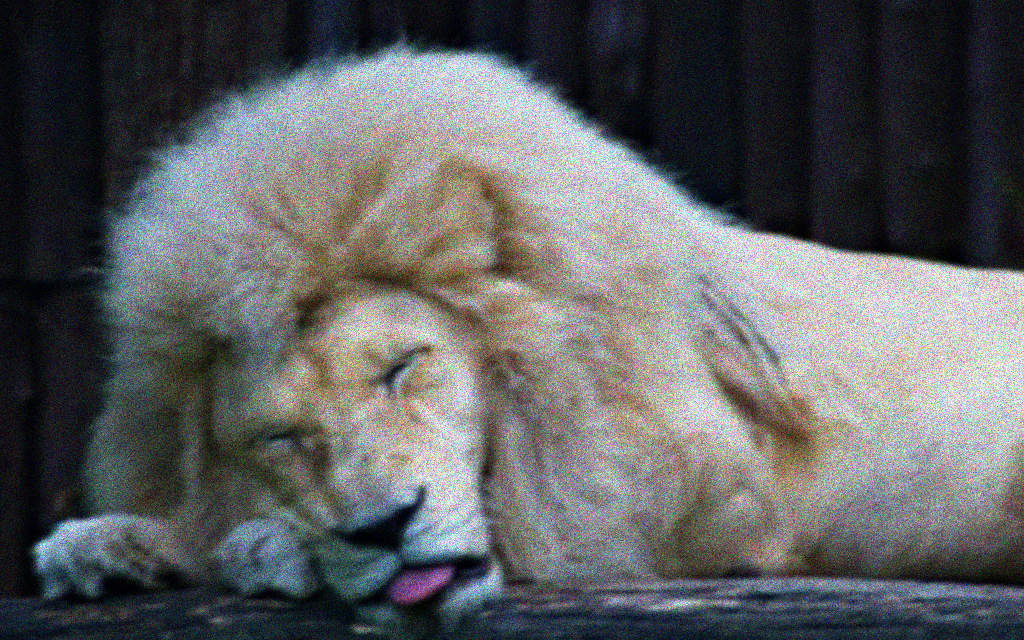}{red}{1.1}{0.9} &
\zoombox{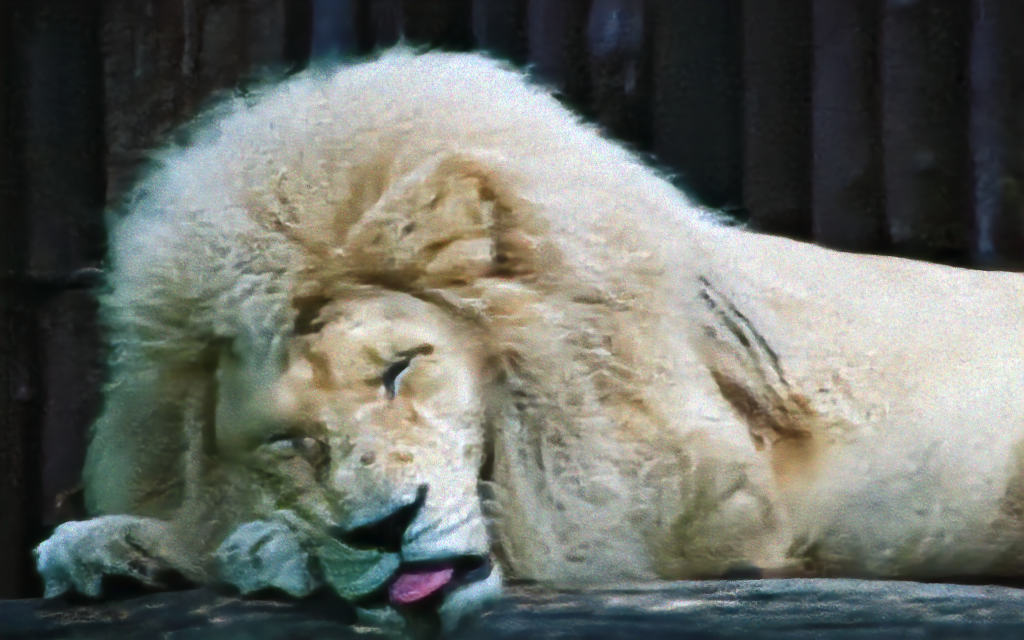}{red}{1.1}{0.9} &
\zoombox{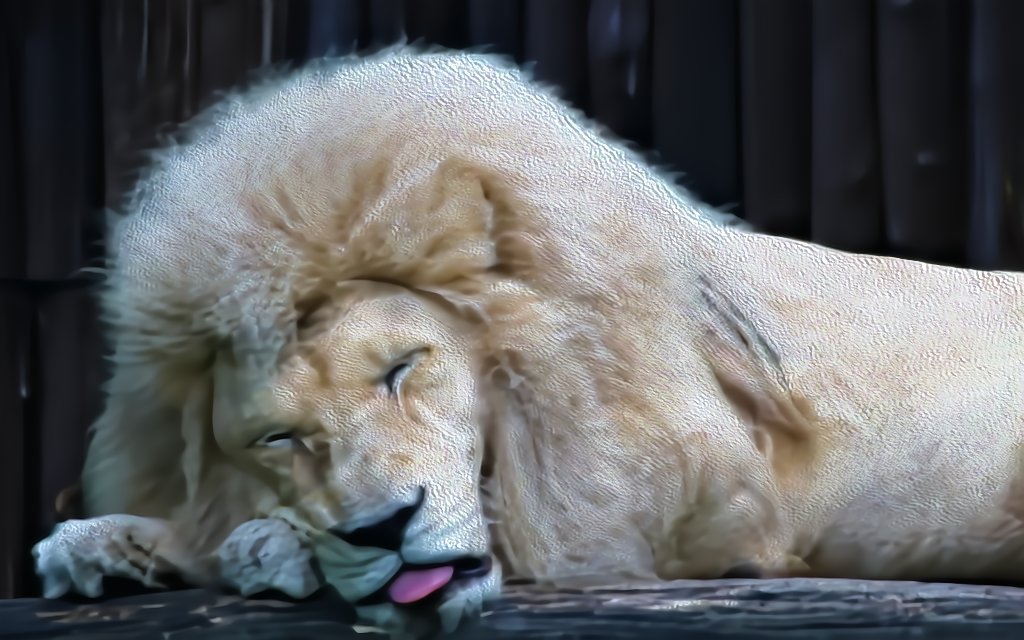}{red}{1.1}{0.9} &
\zoombox{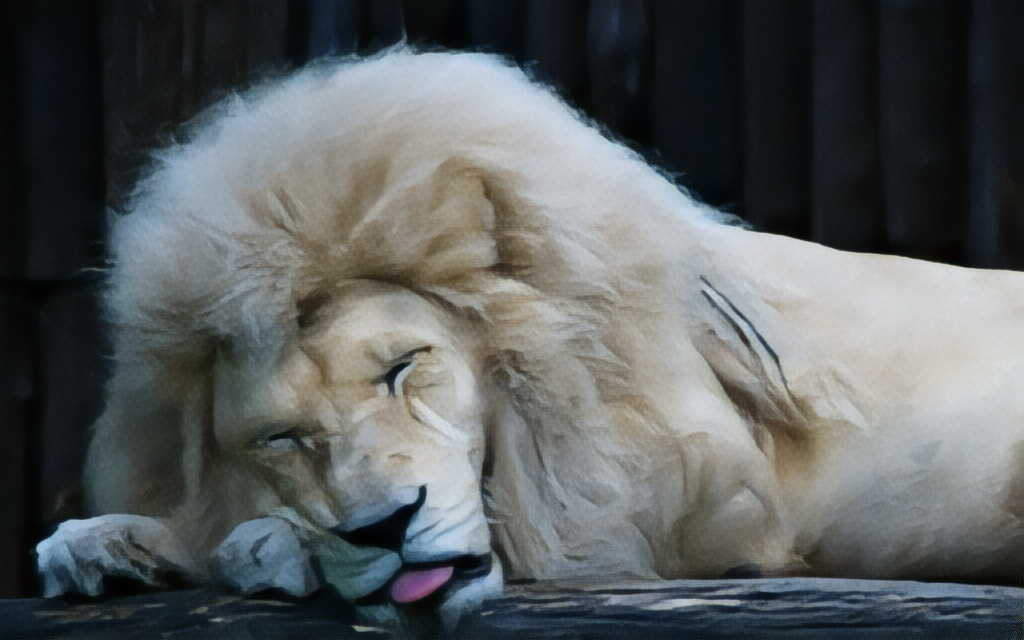}{red}{1.1}{0.9} &
\zoombox{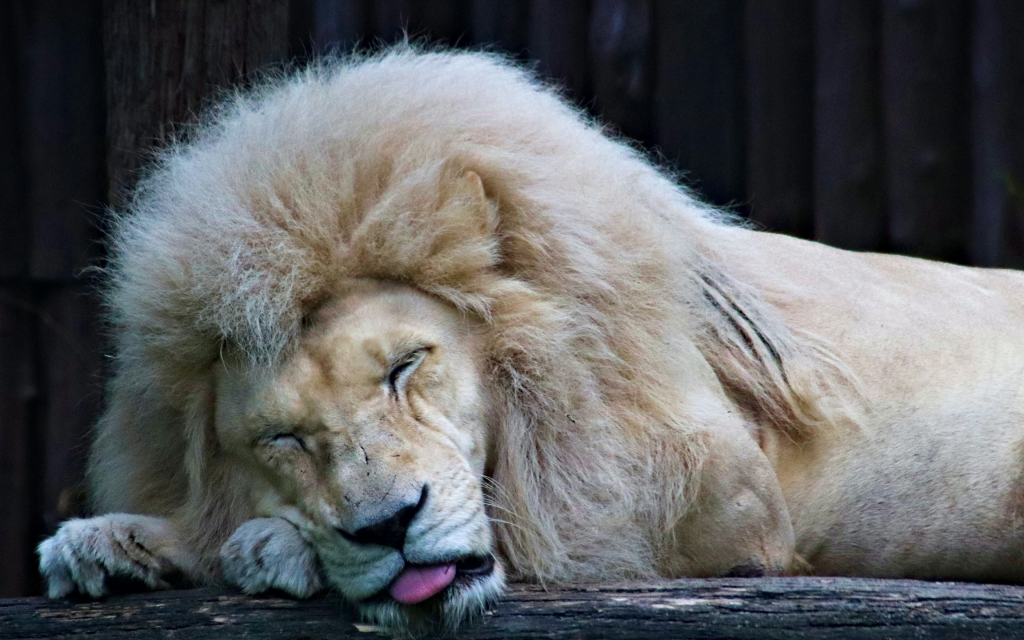}{red}{1.1}{0.9} \\[2pt]
\multicolumn{5}{c}{\small\textit{Rain + Haze}} \\[2pt]
\zoombox{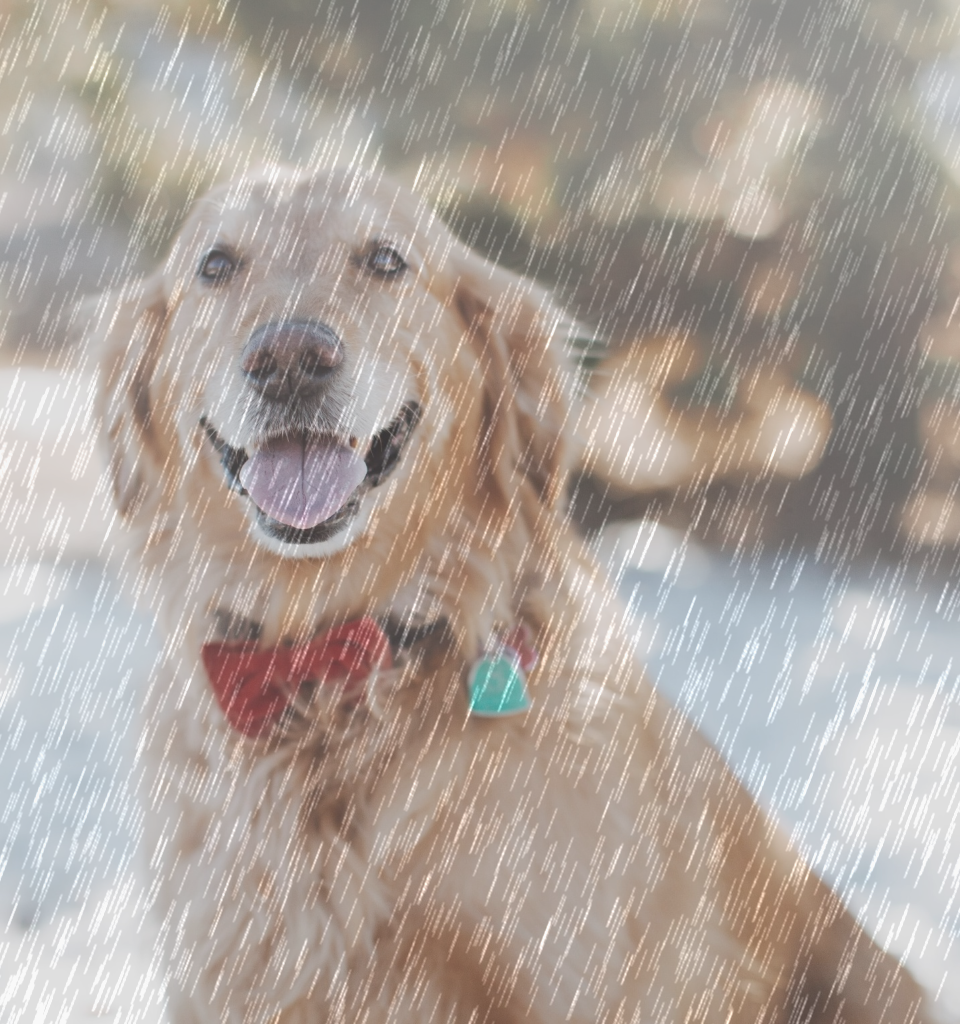}{red}{1.6}{1.6} &
\zoombox{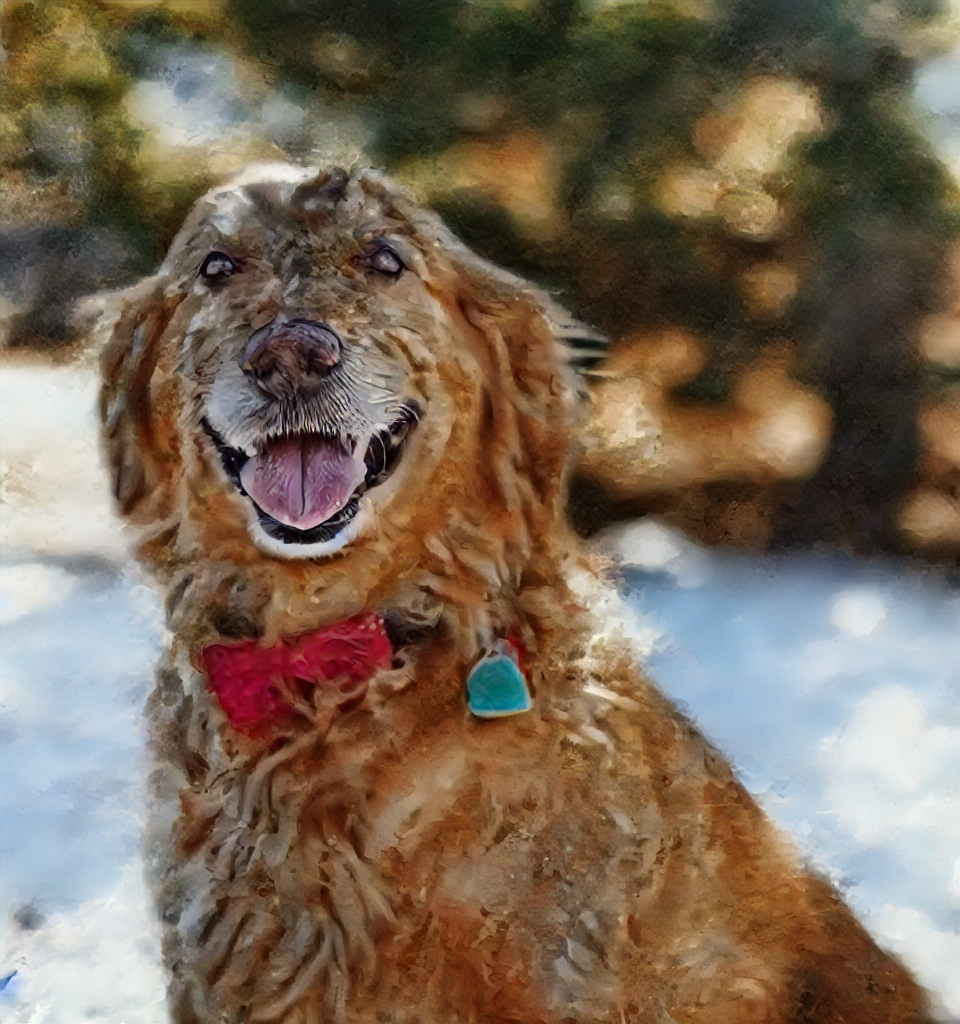}{red}{1.6}{1.6} &
\zoombox{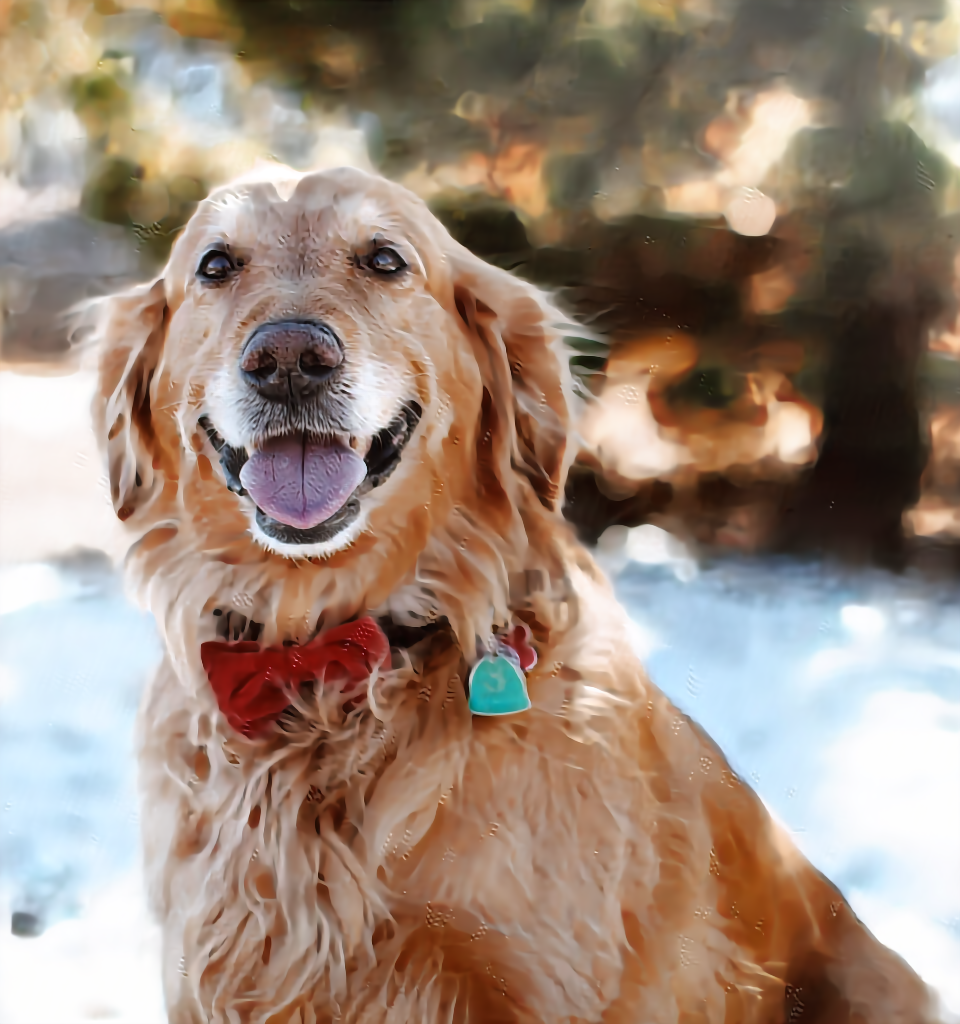}{red}{1.6}{1.6} &
\zoombox{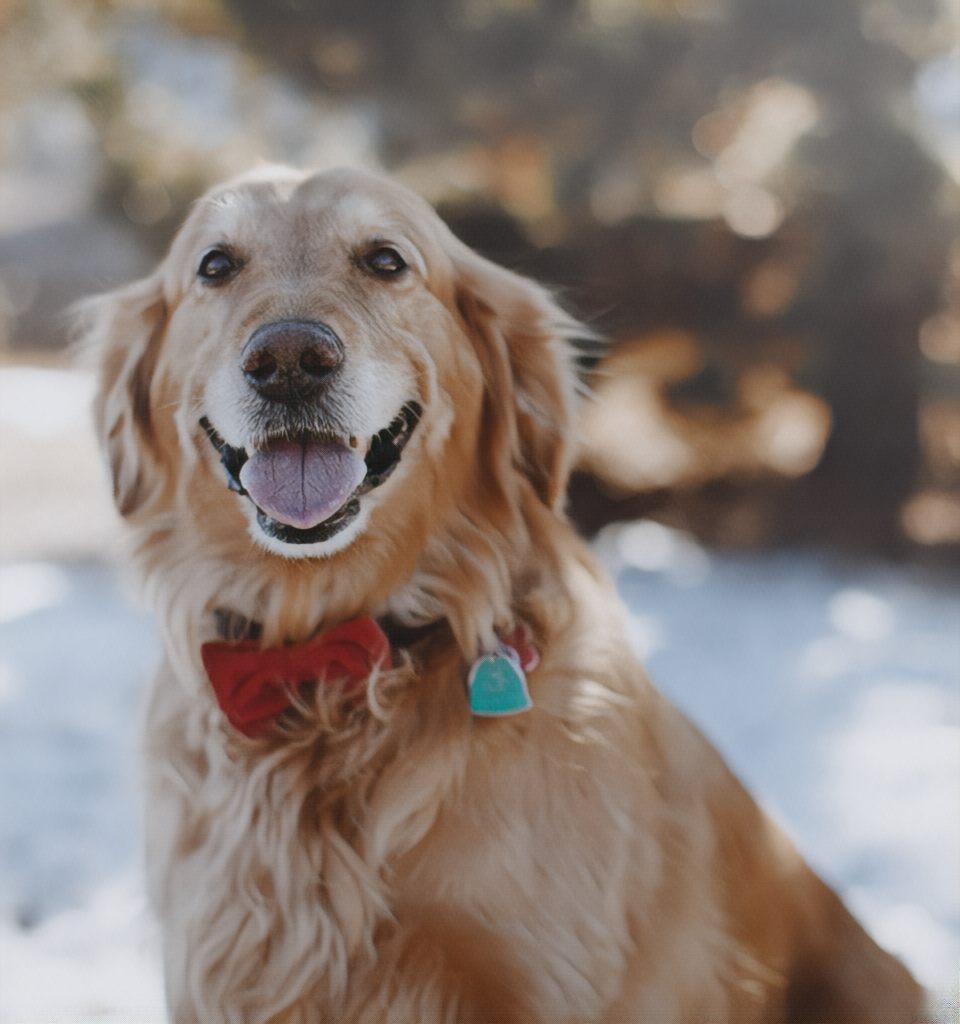}{red}{1.6}{1.6} &
\zoombox{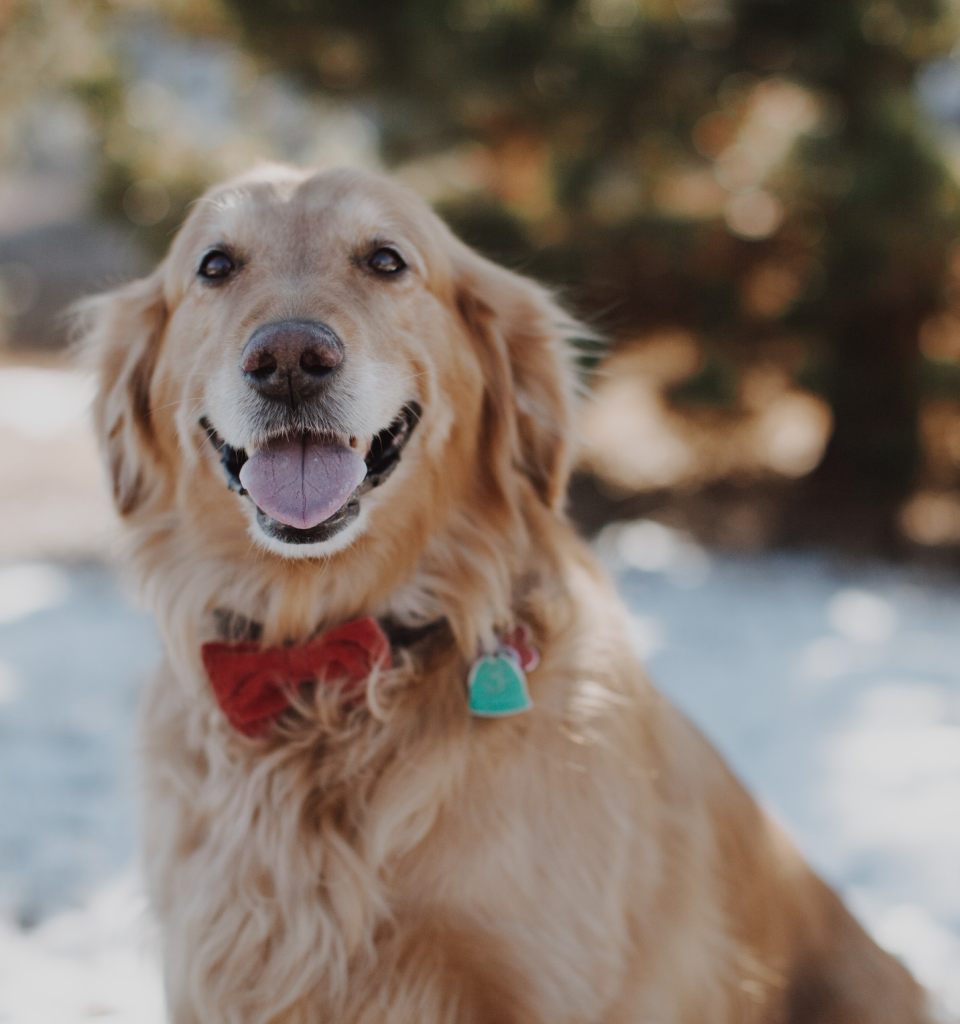}{red}{1.6}{1.6} \\[2pt]
Input & 4KAgent & Ours (Agent) & Ours (Full) & GT \\
\end{tabular}
\caption{Qualitative comparison on benchmarks from AgenticIR~\cite{agenticir}.}
\label{fig:qualitative}

\vskip -0.1in

\end{figure*}
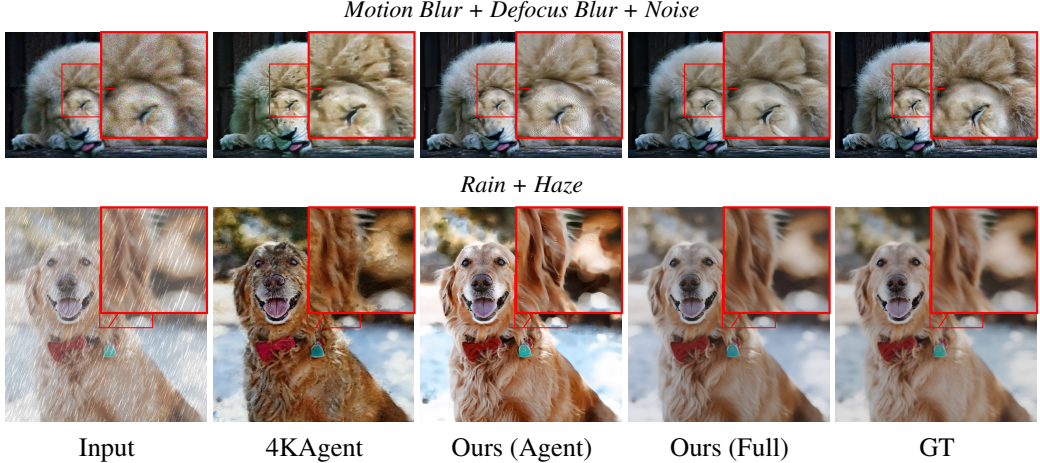

\textbf{Qualitative Comparison.}
\cref{fig:qualitative} presents visual comparisons. Compared to 4KAgent, our method better preserves fine textures such as fur details while effectively removing degradation artifacts.

\subsection{Effectiveness of Planning Agent Optimization}
\label{exp:planning}

In this section, we evaluate whether the proposed training method in~\cref{method_agent_train} effectively equips the model with strong planning capabilities for image restoration.

\textbf{Training Dynamics.} The training process exhibits stable and consistent dynamics, as shown in ~\cref{fig:training_dynamic}. The degradation prediction reward $R_d$ rapidly increases to 0.8 within the first 100 steps, while the restoration reward $R_q$ also improves steadily over time. This suggests that the agent effectively explores the planning space and progressively discovers higher-quality restoration strategies.

\textbf{Behavior Analysis.} We observe that the agent acquires meaningful planning strategies through reinforcement learning, with behaviors that closely align with established domain knowledge. For example, it prioritizes denoising in 89.5\% of noise-corrupted images and follows a derain→dehaze sequence in 75.1\% of cases where both degradations co-occur, consistent with findings in~\cite{agenticir}. The agent also develops more nuanced strategies. Notably, it learns selective repetition: deblurring is frequently applied iteratively (97.3\%), while denoising is rarely repeated (18.7\%), suggesting that the agent identifies which operations benefit from refinement. In addition, the agent tends to invoke more tools than the nominal number of degradations. For images with 1, 2, and 3 degradations, it applies an average of 2.8, 3.8, and 4.4 tools, respectively, in line with observations in~\cref{sec:empirical}.

\subsection{Generalization to Real-World Data}
\label{exp:real_world}
Real-world image degradations are inherently complex, where images often exhibit multiple interacting degradations whose types and boundaries cannot be reliably identified. To evaluate generalization, we test on two widely-used real-world datasets RTTS~\cite{li2018benchmarking} (haze) and LHP~\cite{Guo_2023_ICCV} (rain). For RTTS, we compare against both all-in-one models and task-specific dehazing methods. For LHP, we select baseline methods where the models have not seen the LHP training data, for a fair comparison.

\begin{table*}[t]
\centering

\begin{minipage}{0.528\linewidth}
\centering
\caption{Generalization on RTTS (real haze). }
\label{tab:rtts}
\resizebox{\linewidth}{!}{
\begin{tabular}{llcc}
\toprule
Category & Method & CLIP-IQA$\uparrow$ & MUSIQ$\uparrow$ \\
\midrule
\multirow{5}{*}{All-in-one}
& TransWeather~\cite{valanarasu2022transweather} & 0.292 & 46.27 \\
& DA-CLIP~\cite{daclip} & 0.325 & 53.23 \\
& InstructIR~\cite{Instructir} & 0.370 & 54.46 \\
& WResVLM~\cite{xu2024towards} & 0.371 & 56.09 \\
& PromptIR~\cite{Promptir} & 0.372 & 53.88 \\
\midrule
\multirow{4}{*}{Task-specific}
& KA-Net~\cite{feng2024advancing} & 0.290 & 54.51 \\
& DEA-Net~\cite{chen2024dea} & 0.370 & 54.09 \\
& Deharmer~\cite{guo2022image} & 0.370 & 53.79 \\
& IPC-Dehaze~\cite{fu2025iterative} & \underline{0.440} & \textbf{59.60} \\
\midrule
\multirow{2}{*}{Ours}
& Ours (Agent) & 0.393 & \underline{56.91} \\
& Ours (Full) & \textbf{0.463} & 55.78 \\
\bottomrule
\end{tabular}
}
\end{minipage}
\hfill
\begin{minipage}{0.40\linewidth}
\centering
\caption{Generalization on LHP dataset.}
\label{tab:lhp}
\resizebox{0.95\linewidth}{!}{
\begin{tabular}{lcc}
\toprule
Method & PSNR$\uparrow$ & SSIM$\uparrow$ \\
\midrule
SPANet~\cite{spanet} & 28.00 & 0.8905 \\
PReNet~\cite{prenet} & 27.57 & 0.8595 \\
MPRNet~\cite{mprnet} & 28.41 & 0.8807 \\
GT-Rain~\cite{ba2022gt-rain}  & 28.62 & 0.8675\\
Uformer-B~\cite{wang2022uformer} & 28.74 & \first{0.9262} \\
NeRD-Rain~\cite{chen2024bidirectional} & 30.83 & 0.8854 \\
SCD-Former~\cite{wu2025scdformer} & 29.41 & 0.9127 \\
FAD-Former~\cite{gao2024efficient} &\first{31.27} & 0.8969 \\
\midrule
Ours (Planning) & 25.85 & 0.8301 \\
Ours (Full) & \second{30.93} & \second{0.8995} \\
\bottomrule
\end{tabular}
}
\end{minipage}
\vskip -0.2in
\end{table*}

As shown in~\cref{tab:rtts,tab:lhp}, OPERA generalizes effectively to real-world distributions. Notably, we would like to highlight that current real-world benchmarks are mostly dominated by a single degradation (e.g. rain), which inherently favors task-specific methods over multi-degradation systems like ours. Effective restoration of these datasets typically does not require the coordination of multiple tools. This makes them less suitable for evaluating our method, which focuses on multi-tool cooperative restoration under mixed degradations. Despite this disadvantage, our method achieves competitive or superior performance.

\subsection{Efficiency Analysis}
\label{exp:efficiency}
Our framework requires only a single end-to-end forward pass during inference, where only a single call to the VLM agent is needed. Compared to baseline agent-based methods that rely on heuristic search or greedy exploration over the planning space, our approach avoids repeated trial-and-error executions.
On average, the agent invokes 2.8, 3.8, and 4.4 tools for images containing 1, 2, and 3 degradations, respectively. These tool invocations constitute the only calls to restoration models during inference, resulting in predictable, controllable computational overhead in practice.

%% file: tables/main_metric.tex
\begin{table*}[t]
\centering
\setlength{\tabcolsep}{1pt}
\caption{Quantitative comparison on Group A, B, and C from AgenticIR~\cite{agenticir}. We report full-reference metrics (PSNR$\uparrow$, SSIM$\uparrow$, LPIPS$\downarrow$) and no-reference metrics (MANIQA$\uparrow$, CLIP-IQA$\uparrow$, MUSIQ$\uparrow$). Group A/B/C contain degradation combinations. ``Ours (Planning)'' denotes that only planning is optimized while using pretrained tools. ``Ours (Full)'' includes both planning and execution training. Best results are in \textbf{bold} and second best are \underline{underlined}. Detailed results are shown in~\cref{app:detailed_results}.}
\label{table:quantitative_comparison}
\resizebox{\textwidth}{!}{%
\begin{tabular}{l|cccccc|cccccc|cccccc}
\toprule
\multirow{2}{*}{Method} & \multicolumn{6}{c|}{Group A} & \multicolumn{6}{c|}{Group B} & \multicolumn{6}{c}{Group C} \\
& PSNR & SSIM & LPIPS$\downarrow$ & MANIQA & CLIP-IQA & MUSIQ & PSNR & SSIM & LPIPS$\downarrow$ & MANIQA & CLIP-IQA & MUSIQ & PSNR & SSIM & LPIPS$\downarrow$ & MANIQA & CLIP-IQA & MUSIQ \\
\midrule
\multicolumn{19}{l}{\textit{All-in-One Models}} \\
AirNet & 19.13 & 0.60 & 0.43 & 0.26 & 0.39 & 42.46 & 19.31 & 0.66 & 0.37 & 0.29 & 0.43 & 47.88 & 17.95 & 0.51 & 0.58 & 0.19 & 0.31 & 30.12 \\
PromptIR & 20.06 & 0.61 & 0.41 & 0.26 & 0.40 & 42.62 & 20.47 & 0.67 & 0.34 & 0.29 & 0.43 & 48.10 & 18.51 & 0.52 & 0.58 & 0.19 & 0.31 & 29.71 \\
MiOIR & 20.84 & 0.66 & 0.37 & 0.25 & 0.39 & 47.82 & 20.56 & 0.69 & 0.32 & 0.26 & 0.43 & 51.87 & 15.63 & 0.49 & 0.54 & 0.17 & 0.29 & 37.95 \\
DA-CLIP & 19.58 & 0.60 & 0.43 & 0.24 & 0.41 & 42.51 & 18.56 & 0.59 & 0.44 & 0.24 & 0.42 & 43.70 & 18.53 & 0.53 & 0.53 & 0.19 & 0.35 & 33.87 \\
InstructIR & 18.03 & 0.58 & 0.44 & 0.27 & 0.35 & 45.77 & 18.34 & 0.62 & 0.41 & 0.30 & 0.38 & 50.94 & 17.09 & 0.51 & 0.56 & 0.17 & 0.25 & 33.69 \\
AutoDIR & 19.64 & 0.63 & 0.40 & 0.25 & 0.38 & 47.01 & 19.90 & 0.66 & 0.35 & 0.25 & 0.40 & 49.64 & 18.61 & 0.54 & 0.50 & 0.20 & 0.29 & 37.86 \\
\midrule
\multicolumn{19}{l}{\textit{Agentic Systems}} \\
AgenticIR & 21.04 & 0.68 & 0.31 & 0.31 & 0.45 & 56.88 & 20.55 & 0.70 & 0.31 & 0.32 & 0.46 & 57.57 & 18.82 & 0.55 & 0.45 & 0.27 & 0.39 & 48.68 \\
MAIR & 21.02 & 0.67 & \second{0.30} & 0.33 & 0.48 & 59.19 & 20.92 & 0.70 & \second{0.28} & 0.35 & 0.51 & \second{60.98} & 19.42 & 0.55 & \second{0.41} & \second{0.28} & 0.42 & \second{51.36} \\
4KAgent & 21.48 & 0.67 & 0.30 & \first{0.37} & \second{0.55} & \first{63.19} & 20.95 & 0.67 & 0.30 & \first{0.37} & \second{0.55} & \first{62.69} & 19.77 & 0.56 & 0.43 & \first{0.35} & \second{0.52} & \first{55.56} \\
\midrule
Ours (Planning) & \second{22.32} & \second{0.71} & 0.33 & \second{0.35} & 0.45 & \second{59.34}  & \second{22.28} & \second{0.75} & 0.31  & \second{0.36} & 0.46 & 59.10 & \second{20.75} & \second{0.61} & 0.47 & 0.28 & 0.37 & 49.91 \\
Ours (Full) & \first{24.81} & \first{0.77} & \first{0.23} & 0.32 & \first{0.72} & 57.84  & \first{24.63} & \first{0.78} & \first{0.20} & 0.33 & \first{0.71} & 58.41  & \first{23.04} & \first{0.66} & \first{0.30} & 0.25 & \first{0.72} & 51.27 \\
\bottomrule
\end{tabular}%
}
\end{table*}

%% file: sections/5_conclusion.tex
\section{Conclusion}

In this paper, we introduce OPERA, a unified framework that jointly optimizes agent planning and tool execution for image restoration. OPERA enables the agent to generate complete restoration plans via end-to-end RL while simultaneously adapting tools to cooperate within agent-generated pipelines. Experiments show that end-to-end optimization of both planning and execution enables OPERA to overcome the limitations of existing agentic methods. Despite being trained exclusively on synthetic datasets, OPERA generalizes well to real-world benchmarks. We believe that the insights from OPERA will inform future agent-based image restoration systems, highlighting the importance of joint planning-execution optimization for effective cooperative image restoration.

%% file: sections/6_appendix.tex
\crefalias{section}{appendix}
\crefname{appendix}{Appendix}{Appendices}
\Crefname{appendix}{Appendix}{Appendices}
\newpage
\section{Appendix Overview}

\cref{appendix:empirical} provides detailed experimental settings for the empirical studies presented in~\cref{sec:empirical}.

\cref{appendix:tools} describes the degradations and their corresponding restoration tools used by OPERA.

\cref{app:tool_training} details the implementation of the tool training procedure introduced in~\cref{method_tool_train}.

\cref{app:back_cost} provides a detailed tool training cost analysis.

\cref{app:gpt_compare} compares the trained planning agent with GPT-5.4.

\cref{app:detailed_results} reports detailed quantitative results for all degradation combinations in the test set.

\cref{app:ablation} presents additional ablation studies.

\cref{app:tool_behavior} analyzes how tool training affects tool behavior.

\cref{app:grpo} describes the GRPO algorithm used to train the planning agent.

\cref{app:limitation} discusses the limitations of this paper.

\cref{app:showcase} shows more qualitative comparisons of our method.

\cref{app:prompt} shows the detailed prompt templates.

\section{Empirical Study of Cooperative Multi-Tool Image Restoration}
\label{appendix:empirical}

This section provides implementation details of the empirical study experiment.

\subsection{Restoration Tools}

In this controlled setting, we only consider 4 common degradations, each of which corresponds to one tool, including:
\begin{itemize}
    \item noise: Restormer~\cite{zamir2022restormer} (Trained with $\sigma=50$)
    \item rain: Restormer~\cite{zamir2022restormer}
    \item haze: Restormer~\cite{zamir2022restormer}
    \item blur: X-Restormer~\cite{chen2024comparative}
\end{itemize}

\subsection{Degradation Combinations}
The input degraded images are synthesized by adding the following eight different degradation combinations:
\begin{itemize}
    \item rain, noise
    \item rain, haze
    \item haze, noise
    \item rain, blur
    \item haze, blur
    \item blur, noise
    \item rain, haze, noise
    \item rain, haze, blur
\end{itemize}

\section{Image Restoration Tools}
\label{appendix:tools}
We collect 16 common image restoration models to form the tool pool $\mathcal{M}$. The detailed tool list is shown in \cref{table:tools}. 

We further compare our tool pool with those adopted by baseline agentic methods in \cref{tab:tool_comparison}. Notably, our method relies on a substantially smaller set of tools, which is a strict subset of the tools used by the baseline systems. This observation highlights that the performance improvements stem from the planning capability learned by the planning agent, rather than from access to a larger or more powerful set of tools.

\input{tables/tool_comparison}
\input{tables/tools}

\section{Tool Training Implementation Details}
\label{app:tool_training}

This section provides implementation details for the joint tool optimization framework described in~\cref{method_tool_train}. We detail the progressive loss schedule (\cref{app:loss_schedule}) and loss function formulations (\cref{app:loss_details}).

\begin{figure}[!tb]
  \begin{center}
    \centerline{\includegraphics[width=\columnwidth]{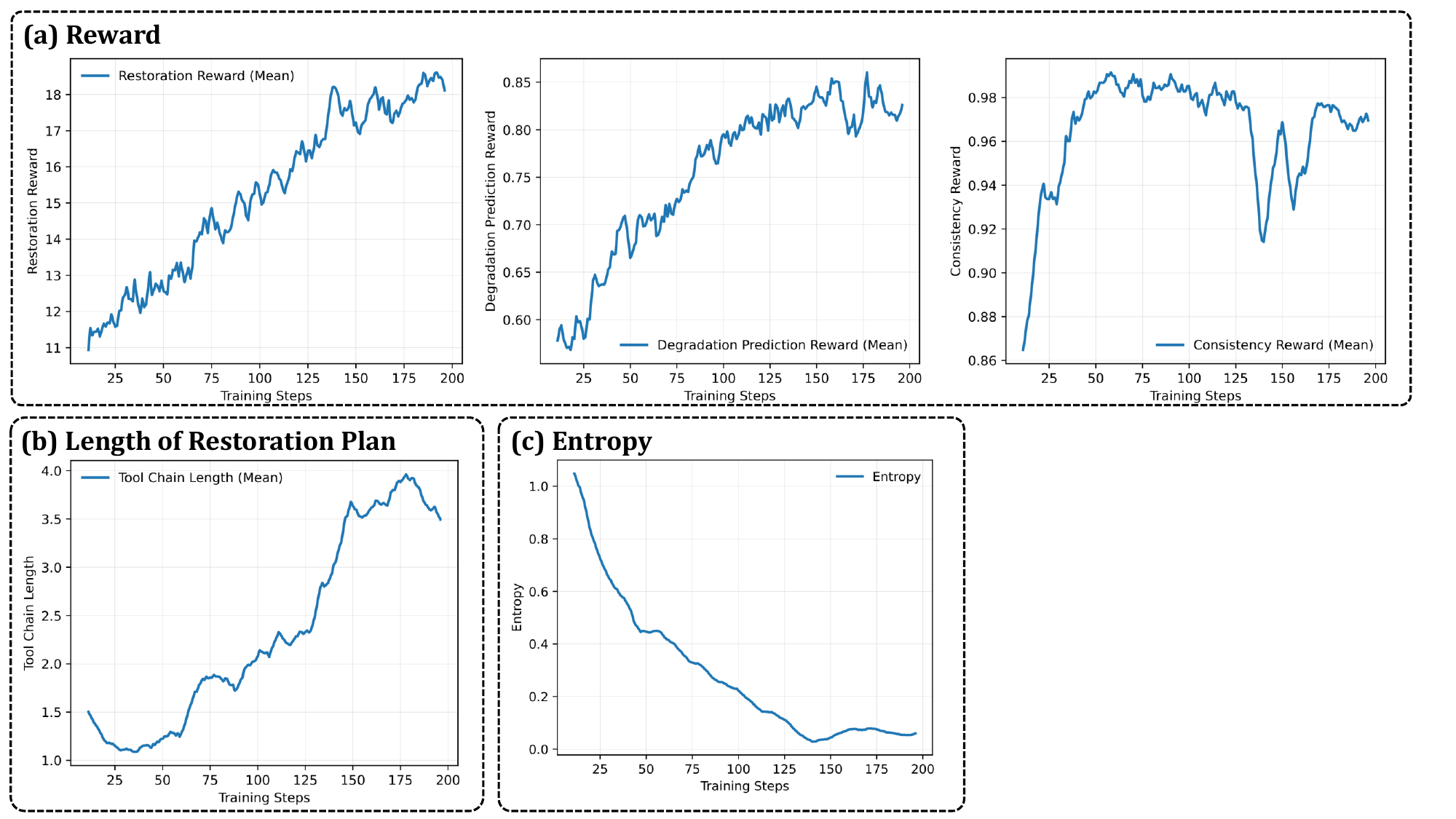}}

    \caption{
The planning optimization GRPO training dynamics.
    }
        \label{fig:training_dynamic}
  \end{center}
  \vskip -0.2in
  
\end{figure}
\subsection{Training Configuration}
\label{app:train_config}

We jointly finetune 15 restoration models from three architectures (Restormer, X-Restormer, SwinIR) covering denoising, deblurring, deraining, dehazing, and super-resolution tasks. Training uses approximately 16,000 synthetic degraded images with an 80/20 train-validation split. We train for 23 epochs with a batch size of 2 and select the checkpoint at epoch 10 based on validation performance.

\textbf{Optimization.} We use the Adam optimizer with learning rate $1 \times 10^{-6}$. Gradient clipping with max norm 0.5 is applied uniformly across all models in each pipeline.

\subsection{Progressive Loss Schedule}
\label{app:loss_schedule}

During tool training, we employ a progressive loss schedule that transitions from pixel-level to perceptual optimization using cosine annealing. This helps stabilize early training when cascaded tools may produce poorly aligned outputs.

Let $e$ denote the current epoch, $E$ the total epochs, and $T = \lfloor 0.3 \cdot E \rfloor$ the transition period. The annealing factor is:
\begin{equation}
\gamma(e) = \frac{1}{2}\left(1 - \cos\left(\pi \cdot \frac{e}{T}\right)\right), \quad e \in [1, T]
\end{equation}

Loss weights evolve as follows during the transition period:
\begin{align}
w_{\text{L1}}(e) &= 1.0 - (1.0 - w_{\text{L1}}^{\text{target}}) \cdot \gamma(e) \\
w_{*}(e) &= w_{*}^{\text{target}} \cdot \gamma(e), \quad * \in \{\text{VGG}, \text{LPIPS}, \text{MUSIQ}, \text{CLIPIQA}\}
\end{align}

After the transition period ($e > T$), all weights remain at their target values. Target weights are: $w_{\text{L1}}^{\text{target}} = 0.4$, $w_{\text{VGG}}^{\text{target}} = 0.1$, $w_{\text{LPIPS}}^{\text{target}} = 0.15$, $w_{\text{MUSIQ}}^{\text{target}} = 0.1$, $w_{\text{CLIPIQA}}^{\text{target}} = 0.1$.

\subsection{Loss Function Details}
\label{app:loss_details}

The total tool training loss combines five components:
\begin{equation}
\mathcal{L} = w_{\text{L1}} \mathcal{L}_{\text{L1}} + w_{\text{VGG}} \mathcal{L}_{\text{VGG}} + w_{\text{LPIPS}} \mathcal{L}_{\text{LPIPS}} + w_{\text{MUSIQ}} \mathcal{L}_{\text{MUSIQ}} + w_{\text{CLIPIQA}} \mathcal{L}_{\text{CLIPIQA}}
\end{equation}

\textbf{L1 Loss} ensures pixel-level fidelity:
\begin{equation}
\mathcal{L}_{\text{L1}} = \frac{1}{N} \sum_{i=1}^{N} |I_{\text{pred}}^{(i)} - I_{\text{gt}}^{(i)}|
\end{equation}

\textbf{VGG Perceptual Loss}~\cite{johnson2016perceptual} captures mid-level feature similarity using VGG19:
\begin{equation}
\mathcal{L}_{\text{VGG}} = \sum_{l \in \mathcal{S}} \frac{1}{C_l H_l W_l} \|\phi_l(I_{\text{pred}}) - \phi_l(I_{\text{gt}})\|_1
\end{equation}
where $\phi_l$ denotes the $l$-th layer features and $\mathcal{S} = \{\text{relu2\_2}, \text{relu3\_4}, \text{relu4\_4}\}$.

\textbf{LPIPS Loss} uses learned perceptual similarity:
\begin{equation}
\mathcal{L}_{\text{LPIPS}} = \sum_{l} \|w_l \odot (\hat{\phi}_l(I_{\text{pred}}) - \hat{\phi}_l(I_{\text{gt}}))\|_2^2
\end{equation}
where $w_l$ are learned channel weights and $\hat{\phi}_l$ are normalized features.

\textbf{MUSIQ Loss} maximizes multi-scale quality score:
\begin{equation}
\mathcal{L}_{\text{MUSIQ}} = 1 - \frac{Q_{\text{MUSIQ}}(I_{\text{pred}})}{100}
\end{equation}

\textbf{CLIP-IQA Loss} maximizes CLIP-based quality assessment:
\begin{equation}
\mathcal{L}_{\text{CLIPIQA}} = 1 - Q_{\text{CLIPIQA}}(I_{\text{pred}})
\end{equation}

The no-reference losses (MUSIQ, CLIP-IQA) encourage realistic appearance without requiring pixel-perfect alignment with ground truth, which is particularly beneficial when multiple valid restoration solutions exist.

\section{Details of Tool Execution Optimization Training Cost}
\label{app:back_cost}
Since the forward and backward passes of execution optimization must traverse the entire active tool chain, we analyze the training cost of tool optimization in this section. 

GPU memory usage is primarily determined by the chain length, batch size, and input resolution. We conduct training on 2$\times$ NVIDIA H20 GPUs (96\,GiB each) using PyTorch DDP, with a per-GPU batch size of 2 and input patches of size $128 \times 128$. The full training process takes approximately two days. In total, 16 tool models are loaded (each with $\sim$26M parameters, corresponding to $\sim$100\,MB per checkpoint).

We log per-iteration GPU memory statistics over 15,720 iterations, summarized in Table~\ref{tab:memory}.

\begin{table}[t]
\centering
\caption{Per-iteration GPU memory usage under different tool chain lengths.}
\label{tab:memory}
\begin{tabular}{c c c c}
\hline
\textbf{Chain Length} & \textbf{Proportion} & \textbf{Forward Memory (GiB)} & \textbf{Peak Reserved (GiB)} \\
\hline
1--2  & 4.2\%  & 7.4  & 18 \\
3--4  & 32.4\% & 16.7 & 26 \\
5     & 35.7\% & 26.2 & 34 \\
6--7  & 26.5\% & 33.5 & 41 \\
8--10 & 1.3\%  & 42.8 & 56 \\
\hline
\end{tabular}
\end{table}

As shown in Table~\ref{tab:memory}, the chain length during training ranges from 1 to 10, with the majority (62\%) falling between 4 and 6. GPU memory usage scales approximately linearly with chain length, increasing by around 5--6\,GiB per additional tool. Given a 96\,GiB GPU memory budget, we estimate the maximum feasible chain length to be approximately 12--14 without further optimization.

In practice, this limit is sufficient. The agent invokes 3--5 tools on average, and even in the most complex cases, the maximum observed chain length is 10, which remains well within the available memory budget.

\section{Comparison of Planning Agent with GPT-5.4.}
\label{app:gpt_compare}
We compare our trained agent against GPT-5.4~\cite{singh2025openai}, which contains strong reasoning and visual understanding capabilities. For a fair comparison, both models are provided with the same tool set and identical prompts. The tools are instantiated by pretrained weights without execution optimization. As shown in~\cref{tab:gpt_compare}, GPT-5.4 demonstrates competitive zero-shot performance. However, our trained agent consistently outperforms GPT-5.4 across all settings and metrics, despite its significantly smaller model size. This demonstrates that while strong proprietary VLMs provide a solid zero-shot baseline, task-specific training remains crucial for achieving robust and high-quality restoration planning, validating the effectiveness of our planning optimization.

\begin{table*}[t]
\centering
\setlength{\tabcolsep}{1pt}
\caption{Quantitative comparison of the planning ability of OPERA and GPT-5.4.}
\label{tab:gpt_compare}
\resizebox{\textwidth}{!}{%
\begin{tabular}{l|cccccc|cccccc|cccccc}
\toprule
\multirow{2}{*}{Method} & \multicolumn{6}{c|}{Group A} & \multicolumn{6}{c|}{Group B} & \multicolumn{6}{c}{Group C} \\
& PSNR & SSIM & LPIPS$\downarrow$ & MANIQA & CLIP-IQA & MUSIQ & PSNR & SSIM & LPIPS$\downarrow$ & MANIQA & CLIP-IQA & MUSIQ & PSNR & SSIM & LPIPS$\downarrow$ & MANIQA & CLIP-IQA & MUSIQ \\
\midrule
Ours (Planning) & \first{21.53} & \first{0.70} & \first{0.32} & \first{0.35} & \first{0.46} & \first{59.02} & \first{21.22} & \first{0.73} & \first{0.31} & \first{0.35} & \first{0.46} & \first{58.54} & \first{20.27} & \first{0.60} & \first{0.46} & \first{0.28} & \first{0.37} & \first{50.62} \\
GPT-5.4 & 21.46 & 0.69 & 0.37 & 0.28 & 0.41 & 50.05 & 20.93 & 0.72 & 0.34 & 0.31 & 0.43 & 52.49 & 20.17 & 0.60 & 0.53 & 0.20 & 0.33 & 36.22 \\

\bottomrule
\end{tabular}%
}
\end{table*}

\section{Detailed Quantitative Results}
\label{app:detailed_results}

\cref{table:per_degradation} provides a detailed breakdown of our full model's performance on each degradation combination in the benchmark used by AgenticIR~\cite{agenticir}. This complements the group-level averages reported in the main text (\cref{table:quantitative_comparison}). 

\input{tables/per_degradation}

\section{Ablation Studies}
\label{app:ablation}

\subsection{Sampling Ratio and Execution Optimization Ablation}
\label{app:ablation_ratio}

We conduct a 2$\times$2 factorial ablation analyzing two factors: (1) degradation sampling ratio (1:1:1 balanced vs.\ 1:3:5 emphasizing multi-degradation), and (2) training scope (planning-only vs.\ full joint planning-execution training). \cref{table:ablation_ratio} presents results on the MiOIR-Test benchmark.

\textbf{Effect of sampling ratio.} Comparing rows with the same training scope, the 1:3:5 ratio consistently outperforms 1:1:1. For agent-only training, MUSIQ improves substantially on Group A, indicating that emphasizing multi-degradation combinations during training improves the agent's tool selection for complex scenarios. Similar trends hold for full training.

\textbf{Effect of joint training.} Comparing planning-only vs.\ full training within each ratio, joint optimization provides consistent improvements across all metrics. Under the 1:1:1 ratio, joint training yields moderate gains. Under the 1:3:5 ratio, the improvements are more pronounced, with perceptual quality metrics showing the largest relative gains. This demonstrates that jointly optimizing tools amplifies the benefits of the sampling strategy.

\textbf{Interaction effect.} The best configuration (1:3:5 + Full) achieves substantially better results than either factor alone, with improvements across all distortion and perceptual metrics compared to the baseline (1:1:1 + Agent only). These results validate our joint training approach: the agent learns better tool composition when tools are simultaneously optimized for multi-degradation cooperation. Based on these results, we adopt the 1:3:5 ratio with full joint training as our final configuration.

\input{tables/ablation_ratio}

\subsection{Reinforcement Learning Reward Ablation}
\begin{figure}[htbp]
    \centering

    \begin{subfigure}[b]{0.3\textwidth}
        \centering
        \includegraphics[width=\linewidth]{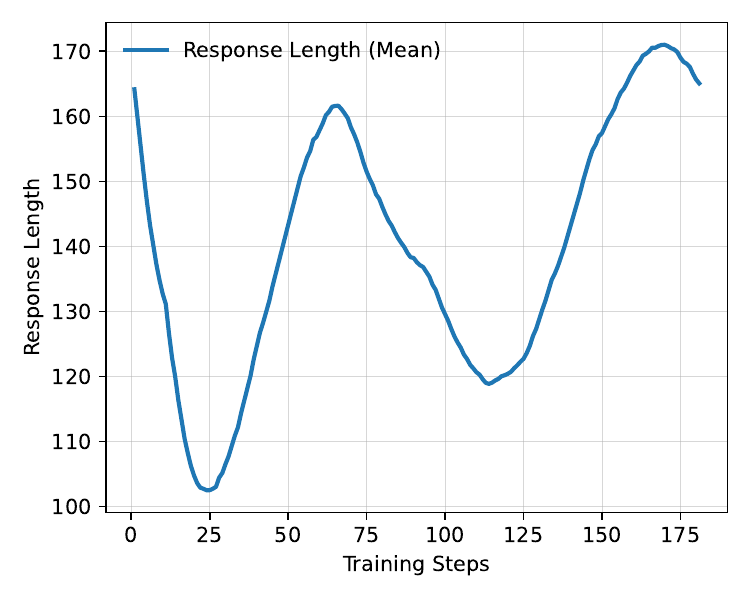}
        \caption{Add Consistency Reward}
        \label{fig:add_consistency}
    \end{subfigure}
    \begin{subfigure}[b]{0.3\textwidth}
        \centering
        \includegraphics[width=\linewidth]{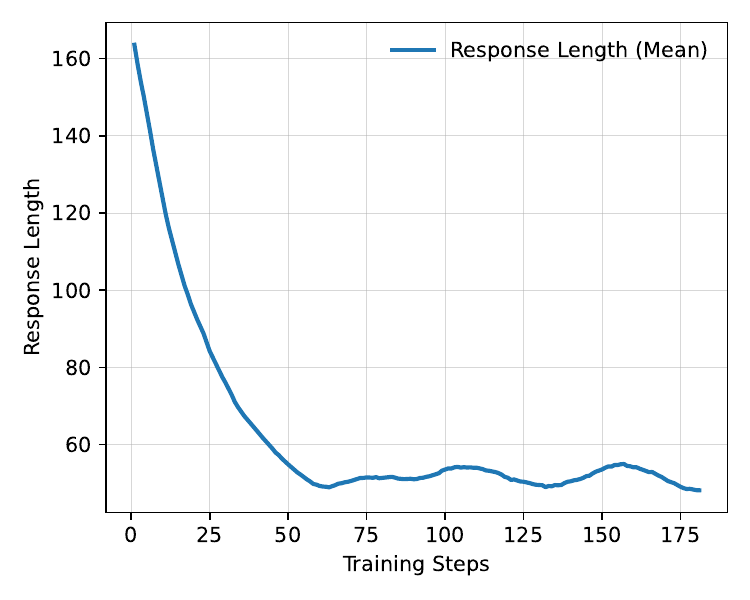}
        \caption{Without Consistency Reward}
        \label{fig:without_consistency}
    \end{subfigure}
    \caption{Comparison of mean response length during training with and without consistency reward.}
    \label{fig:response_length_compare}
\end{figure}

During planning optimization, we introduce a \emph{consistency reward} that evaluates whether the reasoning process aligns with the model's final decision. We employ Pangu-Embedded-7B~\cite{chen2025pangu} to estimate this reward. To assess its effect, we compare the full model with a version where the consistency reward is ablated. As shown in~\cref{fig:response_length_compare}, including the consistency reward prevents the rapid decline in response length, ensuring that the model continues to provide rationales when making decisions.
We further evaluate both versions under the same setting as in~\cref{table:quantitative_comparison}. As shown in~\cref{table:consistency_reward_ablation}, the ablated version consistently underperforms the full model, highlighting the positive impact of the consistency reward.

\begin{table*}[t]
\centering
\setlength{\tabcolsep}{1pt} %

\caption{Ablation study of the consistency reward on Group A, B, and C from AgenticIR~\cite{agenticir}.}
\label{table:consistency_reward_ablation}
\resizebox{\textwidth}{!}{%
\begin{tabular}{l|cccccc|cccccc|cccccc}
\toprule
\multirow{2}{*}{Method} & \multicolumn{6}{c|}{Group A} & \multicolumn{6}{c|}{Group B} & \multicolumn{6}{c}{Group C} \\
& PSNR & SSIM & LPIPS$\downarrow$ & MANIQA & CLIP-IQA & MUSIQ & PSNR & SSIM & LPIPS$\downarrow$ & MANIQA & CLIP-IQA & MUSIQ & PSNR & SSIM & LPIPS$\downarrow$ & MANIQA & CLIP-IQA & MUSIQ \\
\midrule
\multicolumn{19}{l}{\textit{Without Consistency Reward}} \\

Ours (Planning Only) & 21.83 & 0.70 & 0.33 & \textbf{0.36} & 0.47 & 58.96 & 22.03 & 0.76 & 0.31 & 0.35 & 0.46 & \textbf{59.22} & 19.91 & 0.62 & 0.45 & \textbf{0.28} & 0.37 & 50.79 \\
Ours (Full) & 23.26 & 0.76 & 0.24 & 0.31 & 0.70 & 57.54 & 24.61 & \textbf{0.79} & 0.22 & 0.32 & 0.70 & 58.18 & 22.67 & \textbf{0.67} & 0.32 & 0.23 & 0.72 & 48.92 \\
\midrule

\multicolumn{19}{l}{\textit{With Consistency Reward}} \\

Ours (Planning) & 22.32 & 0.71 & 0.33 & 0.35 & 0.45 & \first{59.34}  & 22.28 & 0.75 & 0.31  & \textbf{0.36} & 0.46 & 59.10 & 20.75 & 0.61 & 0.47 & 0.28 & 0.37 & 49.91 \\
Ours (Full) & \first{24.81} & \first{0.77} & \first{0.23} & 0.32 & \first{0.72} & 57.84  & \first{24.63} & 0.78 & \first{0.20} & 0.33 & \first{0.71} & 58.41  & \first{23.04} & 0.66 & \first{0.30} & 0.25 & \first{0.72} & \first{51.27} \\
\bottomrule
\end{tabular}%
}
\end{table*}

\subsection{Ablation Study on Mixed IQA Metrics for Optimization Objective}

In both planning and execution optimization, we adopt an end-to-end training paradigm, where the final image quality serves as the reward/loss. In practice, we optimize a mixture of multiple IQA metrics, as described in~\cref{method_agent_train} and~\cref{method_tool_train}. This section provides further justification for this design.

Image restoration quality is inherently multi-dimensional, involving both pixel-level fidelity and perceptual realism. It is well known that optimizing a single metric (e.g., PSNR alone) often leads to suboptimal perceptual quality. To address this, we adopt a composite objective that combines both full-reference and no-reference IQA metrics, a strategy commonly used in image restoration literature. These metrics provide complementary supervision signals and help mitigate bias toward any single quality aspect. Similar practices have also been adopted in prior agent-based image restoration methods~\cite{chen2024restoreagent, zhou2025q}.

We further conduct an ablation study on planning optimization using a reduced set of metrics (CLIP-IQA and MUSIQ). The results are presented in~\cref{table:iqa_ablation}. The ablated variant shows improved performance on no-reference metrics, but slightly degrades full-reference metrics. This observation highlights the inherent trade-off in objective design, and validates our choice of leveraging multiple complementary metrics to achieve a more balanced overall performance.
\begin{table*}[t]
\centering
\setlength{\tabcolsep}{1pt}
\caption{Ablation study of IQA metrics in planning optimization reward. The ablated version retains only no-reference metrics CLIP-IQA and MUSIQ.}
\label{table:iqa_ablation}
\resizebox{\textwidth}{!}{%
\begin{tabular}{l|cccccc|cccccc|cccccc}
\toprule
\multirow{2}{*}{Method} & \multicolumn{6}{c|}{Group A} & \multicolumn{6}{c|}{Group B} & \multicolumn{6}{c}{Group C} \\
& PSNR & SSIM & LPIPS$\downarrow$ & MANIQA & CLIP-IQA & MUSIQ & PSNR & SSIM & LPIPS$\downarrow$ & MANIQA & CLIP-IQA & MUSIQ & PSNR & SSIM & LPIPS$\downarrow$ & MANIQA & CLIP-IQA & MUSIQ \\
\midrule
Ours (Planning) & \first{21.53} & \first{0.70} & \first{0.32} & 0.35 & 0.46 & 59.02 & \first{21.22} & \first{0.73} & \first{0.31} & 0.35 & 0.46 & 58.54 & \first{20.27} & 0.60 & 0.46 & \first{0.28} & 0.37 & 50.62 \\
Reduced Metrics & 20.86&  0.70&0.33 & \first{0.37} & \first{0.53} & \first{66.29} & 20.16 & 0.70 & 0.35 & \first{0.38} & \first{0.53} & \first{64.65} &  19.83 & \first{0.61} & \first{0.44} & 0.27 & \first{0.47}& \first{61.73}\\

\bottomrule
\end{tabular}%
}
\end{table*}

\section{Tool Behavior Analysis}
\label{app:tool_behavior}

We analyze how tool training affects tool behavior through two experiments.

\textbf{Misuse Experiment.}
Pretrained restoration tools are designed for specific degradations and tend to over-process inputs regardless of whether restoration is necessary. To evaluate whether trained tools exhibit adaptive behavior, we apply each tool to 100 high-quality images containing no degradation and measure the quality degradation introduced.

As shown in \cref{table:tool_behavior} (left), pretrained tools cause noticeable quality degradation when applied to clean images. In contrast, trained tools preserve image quality more effectively, achieving +7.31~dB higher PSNR and 0.073 lower LPIPS. This demonstrates that trained tools better preserve image quality when applied to images that do not require restoration.

\textbf{Single-Degradation Experiment.}
We further examine whether the adaptive behavior observed above comes at the cost of reduced restoration capability. We evaluate each tool on 20 images containing only its corresponding target degradation.

As shown in \cref{table:tool_behavior} (right), trained tools maintain comparable performance to their pretrained counterparts on single-degradation inputs. The minor performance differences suggest that training does not cause catastrophic forgetting of the original restoration capability.

\textbf{Summary.}
These results indicate that trained tools exhibit adaptive behavior, reducing unnecessary modifications while maintaining restoration capability.

\input{tables/tool_behavior}

\section{Group Relative Policy Optimization}
\label{app:grpo}
For reinforcement learning, we adopt Group Relative Policy Optimization (GRPO) \cite{shao2024deepseekmath}, which has been shown to be effective and stable. Given an input $x$ and a VLM policy $\pi_\theta$, GRPO use the reference policy $\pi_{\text{ref}}$ to generate a group of $G$ outputs $\{y_1, y_2, \ldots, y_G\}$. Each response $y_i$ will receive a reward $R_i$. The advantage of each output is calculated by normalizing the rewards within the group: $A_i=\frac{R_i-\text{mean}}{\text{std}}$. The policy $\pi_\theta$ is optimized by the following objective:
\begin{align}
\mathcal{J}_{GRPO}(\theta) = & \mathbb{E}_{\substack{x \sim \mathcal{D}, }} \biggl[ \frac{1}{G} \sum_{i=1}^G \min \biggl(r_{i}(\theta) A_i, 
 \operatorname{clip}\left(r_i(\theta), 1-\varepsilon, 1+\varepsilon\right) A_i\biggr) 
- \beta \mathbb{D}_{KL}(\pi_\theta \| \pi_{\mathrm{0}}) \biggr] 
\end{align}
where $r_i(\theta)=\frac{\pi_\theta(y_i \mid x)}{\pi_{\theta_{\text{old}}}(y_i \mid x)}$, $\varepsilon$ is the clipping parameter, $\beta$ is the KL coefficient.

\section{Limitations and Societal Impacts}
\label{app:limitation}

\textbf{Limitations}. We acknowledge several limitations of our proposed OPERA framework:
\begin{itemize}
\item \textbf{Dependence on the tool pool.} The performance of OPERA is inherently bounded by the coverage, diversity, and quality of the available restoration tools. When certain degradation types are unsupported in the tool pool, the system may fail to fully recover the image, leading to suboptimal restoration results.

\item \textbf{Inference overhead.} Although OPERA is more efficient than search-based agentic systems, it still requires one call to a vision-language model followed by the sequential execution of multiple restoration tools. As a result, the overall inference cost is higher than that of a single end-to-end model, which may limit its applicability in real-time or resource-constrained scenarios.

\item \textbf{Performance trade-offs between generalization and specialization.} Jointly training restoration tools to cooperate improves robustness in complex, multi-degradation settings. However, as shown in our analysis, such joint optimization may slightly degrade the performance of individual tools on single-degradation tasks. This suggests a trade-off between collaborative generalization and task-specific specialization.

\end{itemize}

\textbf{Societal Impacts.} This paper presents a method for advancing image restoration under complex real-world degradations by improving agent-based planning and tool cooperation. Overall, we
believe this work contributes positively to the field of machine learning and computer vision, and its broader societal implications are consistent with established research in image restoration. No significant potential societal consequences of this work must be specifically highlighted here.

\section{More Qualitative Comparisons}
\label{app:showcase}
\begin{figure*}[p]
  \centering
  \includegraphics[width=\textwidth]{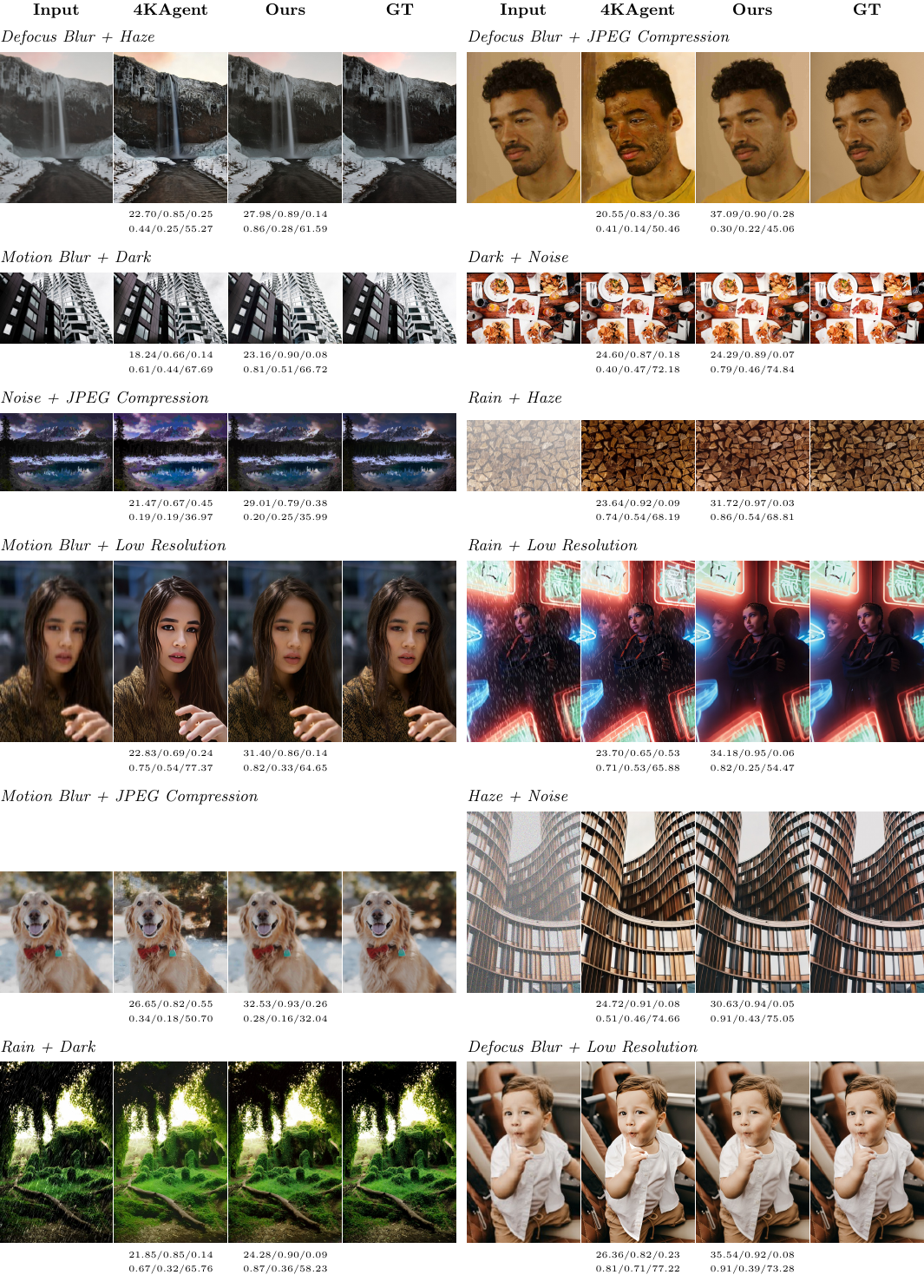}
  \caption{Qualitative comparison across degradation categories
    of Groups~A and~B on the AgenticIR benchmark (part 1/2). Metrics reported: PSNR/SSIM/LPIPS$\downarrow$/MANIQA/CLIP-IQA/MUSIQ.}
  \label{fig:appendix_showcase_p1}
\end{figure*}

\begin{figure*}[t]
  \centering
  \includegraphics[width=\textwidth]{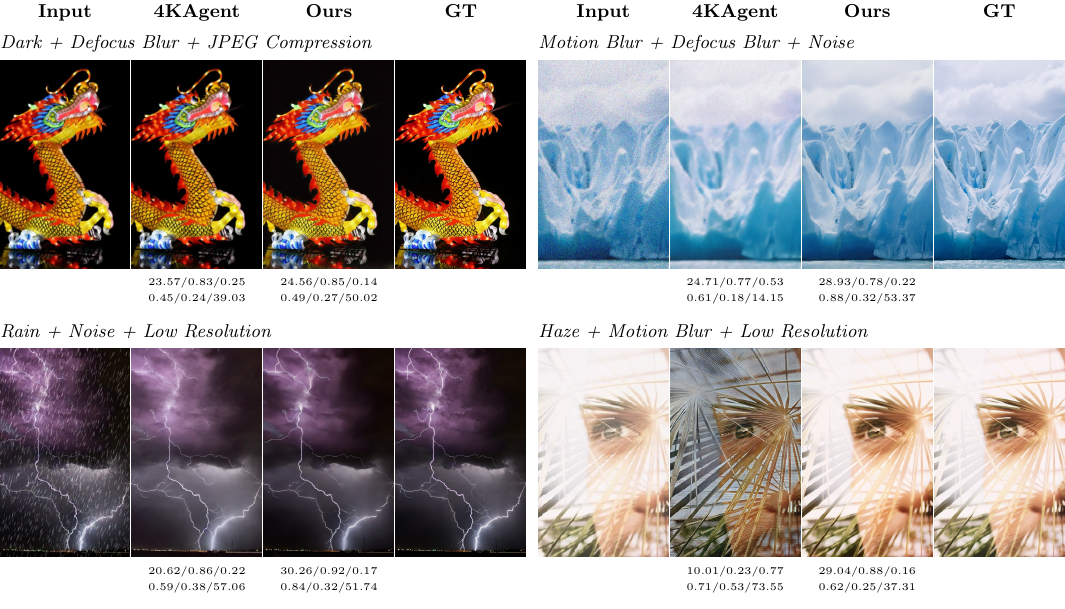}
  \caption{Qualitative comparison on Group~C triple-degradation
    categories (part 2/2).  Metrics reported: PSNR/SSIM/LPIPS$\downarrow$/MANIQA/CLIP-IQA/MUSIQ. Continuation of Fig.~\ref{fig:appendix_showcase_p1}.}
  \label{fig:appendix_showcase_p2}
\end{figure*}

~\cref{fig:appendix_showcase_p1} and ~\cref{fig:appendix_showcase_p2} present more visual comparisons.

\section{Prompt Templates}
\label{app:prompt}
The prompt for the LLM judge to decide the consistency reward is shown in~\cref{tab:consistency_prompt}. The prompt for the planning agent is shown in~\cref{tab:plan_prompt}.

\begin{table}[h]
\centering
\caption{Full prompt used for calculating consistency reward from the LLM judge.}
\label{tab:consistency_prompt}
\begin{tabular}{p{1cm}|p{12cm}}
\toprule
\textbf{Usage} & \textbf{Prompt} \\ \midrule
System Prompt & 
\begin{minipage}[t]{\linewidth}
You are a rigorous planning problem evaluator.
\\

I will provide you with two parts:

A Reasoning Process describing how a planning problem is analyzed and solved

A Final Plan representing the final planning decision or outcome
\\

Your task is to evaluate them according to the following criteria:

1. Evaluate the Reasoning Process

- The reasoning process must NOT be empty

- It must contain meaningful, coherent, and logical reasoning steps

- It should include analysis of constraints, assumptions, or decision logic

- If the reasoning process is missing, empty, superficial, or logically flawed, mark it as unreasonable
\\

2. Check Consistency Between Reasoning Process and Final Plan

- The final plan must be logically derivable from the reasoning process

- There should be no contradictions between the reasoning process and the final plan

- If the reasoning supports one conclusion but the final plan states another, mark them as inconsistent
\\

3. Provide a Clear Judgment and Explanation

Only output a single ``Yes'' or ``No''. Do not provide other explanations or text.
\end{minipage}

\\ \bottomrule
\end{tabular}
\end{table}

\begin{table}[]
\centering
\caption{Full prompt used for planning agent.}
\label{tab:plan_prompt}
\begin{tabular}{p{1cm}|p{12cm}}
\toprule
\textbf{Usage} & \textbf{Prompt} \\ \midrule
System Prompt & 
\begin{minipage}[t]{\linewidth}

You are a professional image restoration assistant.
\\

You will be given an image as input. Your task is to:

1. Visually analyze the image and identify what degradations it contains.

2. Design an optimal sequence of restoration tool calls to enhance the image quality.
\\

\# Possible Degradations:

- noise

- rain

- haze

- defocus\_blur

- motion\_blur

- low\_resolution

- jpeg
\\

\# Tools from Restormer

- restormer.gaussian\_denoise\_15

- restormer.gaussian\_denoise\_25

- restormer.gaussian\_denoise\_50

- restormer.derain

- restormer.defocus\_deblur

- restormer.motion\_deblur
\\

\# Tools from X-Restormer

- xrestormer.denoise\_50

- xrestormer.derain

- xrestormer.dehaze

- xrestormer.deblur

- xrestormer.super\_resolution
\\

\# Tools from SWIN-IR

- swinir.super\_resolution

- swinir.gaussian\_denoise\_15

- swinir.gaussian\_denoise\_25

- swinir.gaussian\_denoise\_50

- swinir.dejpeg
\\

First, think visually in \text{\textless think\textgreater} ~\textless/think\textgreater ~by describing the quality of the image and how you plan to restore it. Then, output a Python list of the detected degradations in \textless degradation\textgreater~ \textless/degradation\textgreater. Finally, output a Python list of the restoration plan in the order in \textless answer\textgreater \textless/answer\textgreater. 
e.g.: \textless think\textgreater thinking progress here \textless /think\textgreater \textless degradation\textgreater['rain', 'noise']\textless/degradation\textgreater \textless answer\textgreater['restormer.gaussian\_denoise\_25', 'xrestormer.derain']\textless/answer\textgreater.
\\

Note that the **order** of tools matters, and tools from different repos may behave differently. Select carefully. 
\end{minipage}

\\ \midrule
User & \textless image\textgreater How to restore this image? Think first then answer.
\\ \bottomrule
\end{tabular}
\end{table}

%% file: tables/tool_comparison.tex
\begin{table}[t]
\centering
\caption{Comparison of tool sets used by different agentic systems. We use a significantly small set of tools, which is a subset of baseline systems.}
\label{tab:tool_comparison}
\begin{tabular}{lcccc}
\toprule
\textbf{Tool} & \textbf{AgenticIR} & \textbf{MAIR} & \textbf{4KAgent} & \textbf{Ours} \\
\midrule
Restormer         & $\checkmark$ & $\checkmark$ & $\checkmark$ & $\checkmark$ \\
X-Restormer       & $\checkmark$ & $\checkmark$ & $\checkmark$ & $\checkmark$ \\
SwinIR            & $\checkmark$ & $\checkmark$ & $\checkmark$ & $\checkmark$ \\
FBCNN             & $\checkmark$ & $\checkmark$ & $\checkmark$ &  \\
DiffBIR           & $\checkmark$ & $\checkmark$ & $\checkmark$ &  \\
DRBNet            &               & $\checkmark$ & $\checkmark$ &  \\
DehazeFormer      & $\checkmark$ & $\checkmark$ & $\checkmark$ &  \\
RIDCP             & $\checkmark$ & $\checkmark$ & $\checkmark$ &  \\
MPRNet            & $\checkmark$ & $\checkmark$ & $\checkmark$ &  \\
MAXIM             &               & $\checkmark$ & $\checkmark$ &  \\
HAT               & $\checkmark$ & $\checkmark$ & $\checkmark$ &  \\
RetinexFormer     &               & $\checkmark$ &               &  \\
DWGAN             &               & $\checkmark$ &               &  \\
CoTF              &               & $\checkmark$ &               &  \\
IFAN              & $\checkmark$ &               & $\checkmark$ &  \\
CLAHE             & $\checkmark$ &               & $\checkmark$ &  \\
NAFNet            &               &               & $\checkmark$ &  \\
ConvIR            &               &               & $\checkmark$ &  \\
LaKDNet           &               &               & $\checkmark$ &  \\
EVSSM             &               &               & $\checkmark$ &  \\
DiffPlugin        &               &               & $\checkmark$ &  \\
FourierDiff       &               &               & $\checkmark$ &  \\
GFPGAN            &               &               & $\checkmark$ &  \\
CodeFormer        &               &               & $\checkmark$ &  \\
\bottomrule
\end{tabular}

\end{table}

%% file: tables/tools.tex
\begin{table}

\centering
\caption{Image restoration models used for different degradation types.}
\begin{tabular}{ll}

\toprule
\textbf{Degradation Type} & \textbf{Model} \\
\midrule

\multirow{7}{*}{Noise}
& Restormer~\cite{zamir2022restormer} (trained with $\sigma=15$) \\
& Restormer~\cite{zamir2022restormer} (trained with $\sigma=25$) \\
& Restormer~\cite{zamir2022restormer} (trained with $\sigma=50$) \\
& X-Restormer~\cite{chen2024comparative} \\
& SwinIR~\cite{liang2021swinir} (trained with $\sigma=15$) \\
& SwinIR~\cite{liang2021swinir} (trained with $\sigma=25$) \\
& SwinIR~\cite{liang2021swinir} (trained with $\sigma=50$) \\
\midrule

\multirow{2}{*}{Rain}
& Restormer~\cite{zamir2022restormer} for deraining \\
& X-Restormer~\cite{chen2024comparative} for deraining \\
\midrule

Dehazing
& X-Restormer~\cite{chen2024comparative} for dehazing \\
\midrule

\multirow{2}{*}{Defocus blur}
& Restormer~\cite{zamir2022restormer} \\
& X-Restormer~\cite{chen2024comparative} \\
\midrule

Motion blur
& Restormer~\cite{zamir2022restormer} \\
\midrule

\multirow{2}{*}{Low-resolution}
& X-Restormer~\cite{chen2024comparative} \\
& SwinIR~\cite{liang2021swinir} \\
\midrule

JPEG Compression Artifact
& SwinIR~\cite{liang2021swinir}\\
\midrule

\multirow{2}{*}{Low Light}
& Constant Shift \\
& Gamma Correction \\

\bottomrule
\end{tabular}

\label{table:tools}

\end{table}

%% file: tables/per_degradation.tex
\begin{table*}[t]
\centering
\caption{Per-degradation performance breakdown of our full model (Ours Full) on the Group A, B, and C from AgenticIR. Results are grouped by degradation complexity (Group A: 2 degradations, Group B: 2 degradations, Group C: 3 degradations).}
\label{table:per_degradation}
\resizebox{\textwidth}{!}{%
\begin{tabular}{llcccccc}
\toprule
Group & Degradation Combination & PSNR$\uparrow$ & SSIM$\uparrow$ & LPIPS$\downarrow$ & CLIP-IQA$\uparrow$ & MUSIQ$\uparrow$ & MANIQA$\uparrow$ \\
\midrule
	\multirow{8}{*}{A}
	& Rain + Haze & 25.25 & 0.94 & 0.06 & 0.84 & 69.07 & 0.44 \\
	& Motion Blur + Low Resolution & 25.78 & 0.72 & 0.18 & 0.84 & 58.78 & 0.28 \\
	& Low Light + Noise & 23.93 & 0.81 & 0.19 & 0.83 & 64.70 & 0.35 \\
	& Defocus Blur + JPEG & 25.75 & 0.68 & 0.43 & 0.34 & 30.69 & 0.19 \\
	& Noise + JPEG & 26.23 & 0.65 & 0.44 & 0.43 & 46.20 & 0.29 \\
	& Rain + Low Resolution & 25.99 & 0.72 & 0.18 & 0.85 & 66.10 & 0.34 \\
	& Motion Blur + Low Light & 22.57 & 0.78 & 0.20 & 0.81 & 60.27 & 0.32 \\
	& Defocus Blur + Haze & 22.98 & 0.82 & 0.14 & 0.86 & 66.88 & 0.34 \\
\midrule
	\multirow{4}{*}{B}
	& Haze + Noise & 22.82 & 0.81 & 0.16 & 0.88 & 66.96 & 0.36 \\
	& Defocus Blur + Low Resolution & 26.74 & 0.73 & 0.17 & 0.86 & 62.85 & 0.31 \\
	& Motion Blur + JPEG & 23.88 & 0.68 & 0.36 & 0.30 & 35.37 & 0.20 \\
	& Rain + Low Light & 25.06 & 0.90 & 0.11 & 0.79 & 68.44 & 0.44 \\
\midrule
	\multirow{4}{*}{C}
	& Haze + Motion Blur + Low Resolution & 21.71 & 0.69 & 0.21 & 0.83 & 56.61 & 0.27 \\
	& Rain + Noise + Low Resolution & 25.56 & 0.70 & 0.22 & 0.85 & 63.62 & 0.31 \\
	& Low Light + Defocus Blur + JPEG & 21.44 & 0.63 & 0.50 & 0.32 & 27.55 & 0.16 \\
	& Motion Blur + Defocus Blur + Noise & 23.46 & 0.61 & 0.27 & 0.87 & 57.30 & 0.27 \\
\midrule
	\multicolumn{2}{l}{Group A Average} & 24.81 & 0.77 & 0.23 & 0.72 & 57.84 & 0.32 \\
	\multicolumn{2}{l}{Group B Average} & 24.63 & 0.78 & 0.20 & 0.71 & 58.41 & 0.33 \\
	\multicolumn{2}{l}{Group C Average} & 23.04 & 0.66 & 0.30 & 0.72 & 51.27 & 0.25 \\
\bottomrule
\end{tabular}%
}
\end{table*}

%% file: tables/ablation_ratio.tex
\begin{table}[t]
\centering
\caption{Ablation study on Group A, B, and C from AgenticIR. We analyze two factors: (1) degradation sampling ratio (1:1:1 balanced vs.\ 1:3:5 emphasizing multi-degradation), and (2) training scope (planning-only vs.\ full optimization). The 1:3:5 ratio combined with joint training yields consistent improvements across all groups.}
\label{table:ablation_ratio}
\begin{minipage}[t]{0.48\linewidth}
\centering
\subcaption{Ratio 1:1:1}
\label{table:ablation_ratio_111}
\resizebox{\linewidth}{!}{
\begin{tabular}{l|l|ccccc}
\toprule
Tool & Group & PSNR$\uparrow$ & SSIM$\uparrow$ & LPIPS$\downarrow$ & CLIPIQA$\uparrow$ & MUSIQ$\uparrow$ \\
\midrule
\multirow{3}{*}{Planning Only} & A & 21.47 & 0.70 & 0.35 & 0.49 & 53.30 \\
 & B & 20.95 & 0.73 & 0.34 & 0.50 & 52.02 \\
 & C & 19.91 & 0.60 & 0.52 & 0.40 & 42.26 \\
\midrule
\multirow{3}{*}{Full} & A & 22.14 & 0.72 & 0.29 & 0.51 & 53.08 \\
 & B & 21.44 & 0.74 & 0.27 & 0.52 & 52.04 \\
 & C & 20.32 & 0.61 & 0.40 & 0.44 & 44.83 \\
\bottomrule
\end{tabular}
}
\end{minipage}
\hfill
\begin{minipage}[t]{0.48\linewidth}
\centering
\subcaption{Ratio 1:3:5}
\label{table:ablation_ratio_135}
\resizebox{\linewidth}{!}{
\begin{tabular}{l|l|ccccc}
\toprule
Tool & Group & PSNR$\uparrow$ & SSIM$\uparrow$ & LPIPS$\downarrow$ & CLIPIQA$\uparrow$ & MUSIQ$\uparrow$ \\
\midrule
\multirow{3}{*}{Planning Only} & A & 21.53 & 0.70 & 0.32 & 0.46 & \textbf{59.02} \\
 & B & 21.22 & 0.73 & 0.31 & 0.46 & \textbf{58.54} \\
 & C & 20.27 & 0.60 & 0.46 & 0.37 & 50.62 \\
\midrule
\multirow{3}{*}{Full} & A & \textbf{23.26} & \textbf{0.76} & \textbf{0.22} & \textbf{0.72} & 57.65 \\
 & B & \textbf{22.81} & \textbf{0.76} & \textbf{0.20} & \textbf{0.70} & 57.96 \\
 & C & \textbf{21.85} & \textbf{0.64} & \textbf{0.30} & \textbf{0.72} & \textbf{51.94} \\
\bottomrule
\end{tabular}
}
\end{minipage}
\end{table}

%% file: tables/tool_behavior.tex
\begin{table}[t]
\centering
\caption{Tool behavior analysis. \textbf{Left}: Misuse experiment evaluates tools on high-quality images without degradation. Trained tools substantially reduce quality degradation, indicating adaptive behavior. \textbf{Right}: Single-degradation experiment evaluates tools on images with only their target degradation. Trained tools maintain comparable performance, confirming no catastrophic forgetting.}
\label{table:tool_behavior}
\begin{minipage}[t]{0.48\linewidth}
\centering
\subcaption{Misuse (HQ images)}
\label{table:misuse}
\resizebox{\linewidth}{!}{
\begin{tabular}{l|ccc}
\toprule
Setup & PSNR$\uparrow$ & SSIM$\uparrow$ & LPIPS$\downarrow$ \\
\midrule
HQ Reference & $\infty$ & 1.000 & 0.000 \\
Pretrained & 40.80 & 0.938 & 0.084 \\
Trained & \textbf{48.11} & \textbf{0.990} & \textbf{0.011} \\
\midrule
$\Delta$ (Trained - Pretrained) & +7.31 & +0.052 & -0.073 \\
\bottomrule
\end{tabular}
}
\end{minipage}
\hfill
\begin{minipage}[t]{0.48\linewidth}
\centering
\subcaption{Single-degradation}
\label{table:single}
\resizebox{\linewidth}{!}{
\begin{tabular}{l|ccc}
\toprule
Setup & PSNR$\uparrow$ & SSIM$\uparrow$ & LPIPS$\downarrow$ \\
\midrule
Pretrained & \textbf{28.62} & \textbf{0.767} & \textbf{0.282} \\
Trained & 28.50 & 0.739 & 0.317 \\
\midrule
$\Delta$ (Trained - Pretrained) & -0.12 & -0.028 & +0.035 \\
\bottomrule
\end{tabular}
}
\end{minipage}
\end{table}